\documentclass{article}
\usepackage[main,final]{neurips_2026}
\usepackage[utf8]{inputenc}
\usepackage[T1]{fontenc}
\usepackage{hyperref}
\usepackage{url}
\usepackage{booktabs}
\usepackage{amsfonts}
\usepackage{nicefrac}
\usepackage{microtype}
\usepackage{xcolor}
\usepackage{colortbl}
\usepackage{amsmath}
\usepackage{enumitem}
\usepackage{longtable}
\usepackage{array}
\usepackage{multirow}
\usepackage{graphicx}
\usepackage{float}
\usepackage{tikz}
\usetikzlibrary{arrows.meta, positioning}
\setcitestyle{numbers}
\title{Knowing When to Ask: Segment-Level Credit Assignment for LLM Tool Use}
\author{
  Abhijit Kumar\thanks{Corresponding author.} \quad Zoey Wu \quad Mohit Suley\thanks{Work done while at Microsoft AI.} \\
  Microsoft AI \\
  Redmond, USA \\
  \texttt{\{abhijitkumar, guwu\}@microsoft.com}
}
\begin{document}
\maketitle
\begin{center}
\url{https://aka.ms/know-when-to-ask}
\end{center}

\begin{abstract}
Humans know when to reach for help: $347 \times 28$ warrants a calculator while $2 + 2$ does not. Language models do not do this by default. Prompt-based approaches can instruct a model when to invoke tools but do not teach the model to decide when it needs tools based on its own knowledge. Reinforcement learning approaches that assign a single outcome reward to the whole trajectory fare no better: trajectory-level credit cannot isolate which tool call in a successful episode actually helped or penalize unnecessary calls. We propose \textbf{CARL} (\textbf{C}ompetence-\textbf{A}ware \textbf{R}einforcement \textbf{L}earning), which trains a critic on the model's own rollouts to learn where the model's parametric knowledge suffices and where it needs external help. By decomposing each rollout at natural tool-use boundaries (e.g., code-fence delimiters and context-block transitions), CARL assigns independent credit to each segment from a single binary outcome, without external judges or step-level annotations, addressing the credit-assignment limitations of trajectory-level methods. As a result, erroneous tool calls, incorrect extractions and unnecessary calls each receive appropriately signed advantages. We show that the trained critic captures the model's domain competence: it separates parametrically solvable from tool-dependent questions with AUC 0.93 at 7B. On five benchmarks spanning arithmetic, multi-hop factual QA and numerical reasoning over financial tables, CARL improves average exact-match accuracy on the three search-based benchmarks by 6.7 points at 7B and 9.6 points at 3B over the best RL baseline (Search-R1 PPO at 7B, Search-R1 GRPO at 3B), with the largest gain (+8.3 EM at 7B, +12.5 EM at 3B) on MuSiQue, the most compositional multi-hop benchmark. At 7B, CARL has a 53\% lower tool-use rate than Search-R1 PPO on questions answerable from parametric knowledge while remaining 9.8 EM points more accurate on those same questions, an emergent consequence of the critic learning where this model's competence ends. Gains are largest at small scale: the 3B model improvement is $1.4\times$ the 7B improvement, suggesting that knowing when to ask for help disproportionately benefits models with smaller parametric memory.
\end{abstract}
\section{Introduction}
\label{sec:intro}

Out of the box, large language models do not come equipped with the machinery that makes their outputs reliable. They lack reliable access to current information, arithmetic precision and a means of verifying their own claims. Prompt-based techniques such as retrieval-augmented generation, ReAct and tool-augmented prompting typically bolt a calculator and a search box onto the model at inference time, yet the model still invokes tools without knowing whether the task actually requires them. Humans do not behave this way. We do not need to be told that $347 \times 28$ warrants a calculator while $2 + 2$ does not, or that the capital of an unfamiliar country requires a search while the capital of our own does not. We recognize the edge of our own knowledge and reach for help precisely when we need it. Our goal is to train language models to do the same, not by being instructed when to use tools, but by learning it.

Recent RL approaches train LLMs on tool-use trajectories with a single outcome reward~\citep{jin2025searchr1,li2025torl,qian2025toolrl,chen2025research,li2025r3rag}, creating two coupled problems: credit assignment (a good search and a bad search in the same successful trajectory are reinforced identically) and tool-use selectivity (the model lacks awareness of its own competence boundary, so unnecessary tool calls still receive positive reinforcement). Some approaches add tool-specific rewards but retain trajectory-level credit. Others score individual steps with an external judge whose assessment may not reflect the model's own knowledge (\S\ref{sec:rw}).

Our starting observation is that tool-use trajectories have something pure reasoning trajectories lack: \emph{structurally observable boundaries} (\S\ref{sec:method:segments}, Figure~\ref{fig:pipeline}). At each boundary, the context changes observably, making a central RL problem tractable: identifying meaningful decision boundaries in generation, a challenge discussed at length in prior work~\citep{kazemnejad2024vineppo,setlur2025rewarding,kumar2026egca}. A critic estimates the probability of a correct final answer from the context at each boundary. We train it using only binary outcome rewards without step-level annotations or boundary heuristics. Given such a critic, the advantage of a segment is simply how much it changed the expected probability of success: a good search raises it, a bad search lowers it, and an unnecessary search on an easy question leaves it roughly unchanged. Because the critic is trained on \emph{this model's own rollouts}, it learns a competence-aware value function, one that reflects what this model can do.

We instantiate this as CARL (\textbf{C}ompetence-\textbf{A}ware \textbf{R}einforcement \textbf{L}earning). Our contributions:

\begin{enumerate}[leftmargin=*,itemsep=4pt]

\item \textbf{A value function that encodes the model's own competence boundary.} From binary outcome reward alone, $V(s_0)$ learns to separate parametrically solvable from tool-dependent questions (AUC 0.93 at 7B, 0.85 at 3B), producing 53\% fewer unnecessary tool calls than the best RL baseline.

\item \textbf{Per-segment credit from a single outcome.} We derive per-segment advantages from the Semi-Markov Decision Process (SMDP)~\citep{sutton1999options} Bellman equation at structurally observable tool-use boundaries. A useful search and a faulty extraction can receive opposing advantages within the same trajectory rather than sharing one trajectory-wide advantage.

\end{enumerate}

\noindent On five benchmarks, CARL improves over Search-R1 PPO on the three search-based datasets, with the largest gains on multi-hop out-of-distribution tasks and up to 36\% fewer tokens. Gains are largest at small scale ($1.4\times$ at 3B vs.\ 7B).
\section{Related work}
\label{sec:rw}
\label{sec:options_smdp}
\label{sec:rw:credit}

Three lines of work are most relevant. Prompt-based methods (RAG, ReAct) attach tools at inference time but cannot train selectivity. Outcome-reward RL methods train from final answers, with trajectory-, token- or segment-level credit. Some step-level methods use additional supervision or external judges whose knowledge may diverge from the student's. We position CARL against all three, with Search-R1 PPO and GRPO as our direct experimental baselines.

\paragraph{Trajectory-level RL for tool use.}
Search-R1~\citep{jin2025searchr1}, ToRL~\citep{li2025torl}, ReSearch~\citep{chen2025research} and R3-RAG~\citep{li2025r3rag} train LLMs on tool-use trajectories with a single outcome reward. Because the reward is shared across all segments, the contribution of any single tool call is hard to isolate and unnecessary tool calls on easy questions still receive positive reinforcement when the episode succeeds. ToolRL~\citep{qian2025toolrl} enriches the reward with tool-name and argument-quality components but applies the result as a single trajectory-level scalar, so it refines what the reward evaluates without changing which segment each signal reaches. StepTool~\citep{yu2024steptool} uses GPT-4 to score each step during PPO. Its feedback reflects the judge's assessment rather than the model's own competence. The two approaches are orthogonal: a richer ToolRL-style reward could in principle be combined with CARL's segment-level credit assignment.

\paragraph{Options, SMDPs and credit assignment.}
CARL instantiates the Options framework~\citep{sutton1999options}: each segment is an option, termination fires at a structural delimiter and the SMDP Bellman equation gives the per-segment advantage (\S\ref{sec:method:advantage}). SPO~\citep{guo2025spo} independently arrives at segment-level credit but uses arbitrary cutpoints and critic-free MC estimation on single-turn math; CARL's boundaries are SMDP environment transitions and its critic encodes a cross-question competence boundary (Appendix~\ref{app:extended_rw}). Recent work applies Options vocabulary to LLM agents (ARC~\citep{taparia2026arc}, Agent-as-Tool~\citep{zhang2025agentastool}) but trains with trajectory-level GRPO; CARL derives per-option advantages from a value function. HICRA~\citep{wang2025hicra} addresses credit assignment differently, weighting optimization toward planning tokens identified through Strategic Gram phrase matching. Among credit-assignment methods, EGCA~\citep{kumar2026egca} localizes credit in code generation via execution-trace comparison but requires reference solutions; PAVs~\citep{setlur2025rewarding} share our $V(s_{t+1}) - V(s_t)$ form but require step-level annotation; VinePPO~\citep{kazemnejad2024vineppo} found token-level value networks unreliable, a problem our segment-level setting avoids because boundaries number in single digits and each corresponds to a large context change. Earlier return-decomposition work (RUDDER~\citep{arjona2019rudder}) shares the potential-based form of our advantages but operates at the token level. SMART~\citep{kim2025smart} uses selective larger-model assistance, while PiCA~\citep{liu2026pica} uses pivot-supervised search credit; detailed comparisons are in Appendix~\ref{app:extended_rw}.

\section{Methodology}
\label{sec:method}

CARL has three components: a decomposition of tool-use trajectories into segments with structurally observable boundaries (\S\ref{sec:method:segments}), a value-function-based advantage derived from the SMDP Bellman equation (\S\ref{sec:method:advantage}) and a critic trained on outcome reward with curated warm-up data (\S\ref{sec:method:critic}). These plug into standard PPO~\citep{schulman2017ppo} (\S\ref{sec:method:ppo}).

\subsection{Segment decomposition and tool interface}
\label{sec:method:segments}

\begin{figure}[t]
\centering
\includegraphics[width=\textwidth]{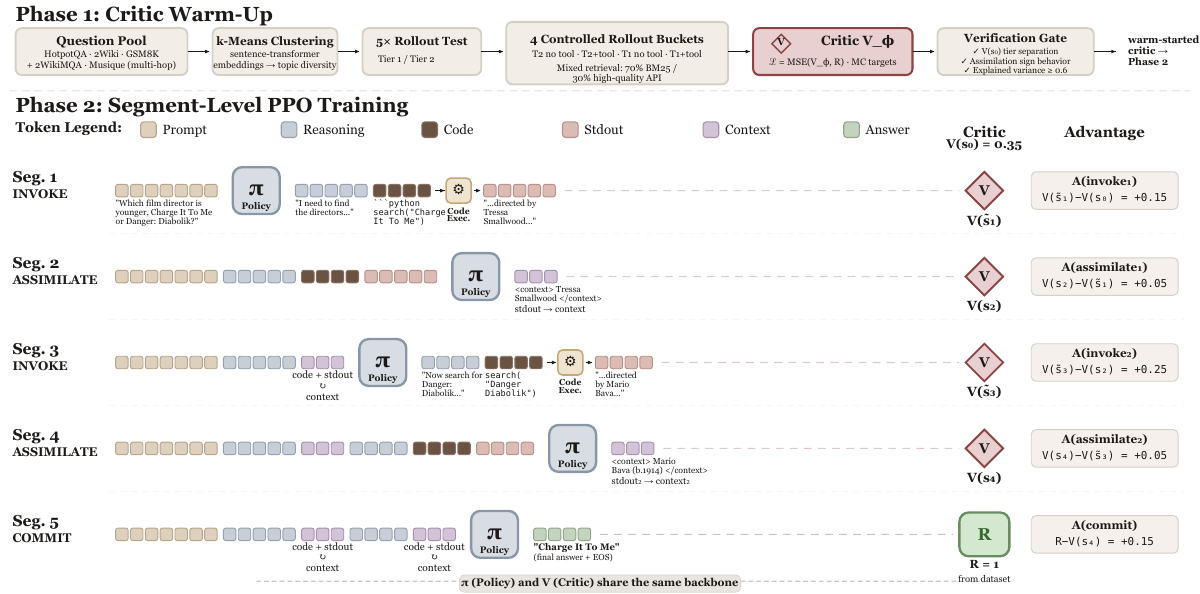}
\caption{\textbf{Rollout pipeline in CARL.}
Each tool-use trajectory decomposes into three segment types (\textbf{invoke}, \textbf{assimilate}, \textbf{commit}) with structurally observable boundaries. After invoke, the rollout loop executes the code and captures stdout; after assimilate, it replaces the raw stdout with the model's \texttt{<context>} block. The critic $V_\phi$ evaluates the context at each boundary; the advantage of each segment is the change in $V_\phi$ across it, with the terminal reward $R$ replacing the final $V_\phi$ at EOS. MuSiQue is excluded from training, MuSiQue example shown is for illustration only.}
\label{fig:pipeline}
\end{figure}

A tool-use trajectory consists of a sequence of segments, each an option in the sense of \citet{sutton1999options}. A trajectory with $K$ tool calls has $2K + 1$ segments; a trajectory with no tool calls is a single commit segment. The tool interface is Python code blocks: the model writes arithmetic, calls a predefined \texttt{search()} function, or implements custom logic. Figure~\ref{fig:pipeline} shows the rollout control flow.

The \emph{invoke} segment generates reasoning followed by a code block (terminated at the closing code fence); the rollout loop executes the code and appends stdout $o_k^{\text{raw}}$. The \emph{assimilate} segment reads tool output and writes a \texttt{<context>} block distilling the relevant information (terminated at \texttt{</context>}); the rollout loop replaces $o_k^{\text{raw}}$ with the extraction $e_{k+1}$. The \emph{commit} segment produces the final answer (terminated at EOS). Because raw tool output is replaced before the next generation begins, each segment's target contains only model-produced tokens and gradient masking is unnecessary (formalized in Appendix~\ref{app:posmdp}).

\paragraph{Reward.}
A single terminal reward $R = \text{EM}(\text{prediction}, \text{gold}) \in \{0,1\}$ after standard normalization. No intermediate rewards, execution penalties, or format rewards.

\subsection{Segment-level advantage}
\label{sec:method:advantage}

Given only a terminal reward $R \in \{0,1\}$, how do we assign credit per segment? From the SMDP Bellman equation (Appendix~\ref{app:smdp_derivation}), under terminal-only reward and finite horizon ($\gamma = 1$), the per-segment advantage reduces to:
\begin{equation}
A(\text{seg}_k) =
\begin{cases}
V_\phi(s_{k+1}) - V_\phi(s_k), & k < N \quad\text{(intermediate segment)},\\[2pt]
R - V_\phi(s_N), & k = N \quad\text{(final segment)}.
\end{cases}
\label{eq:segment_advantage}
\end{equation}
For a single tool-interaction trajectory with segments (invoke, assimilate, commit), this yields three independent advantages from one binary outcome. Summing per-segment advantages telescopes to $R - V_\phi(s_0)$, preserving the same total credit as trajectory-level PPO (Appendix~\ref{app:smdp_derivation}).

The same equations apply to graded terminal rewards in $[0,1]$; Appendix~\ref{app:graded-reward} evaluates this setting.

\paragraph{Per-segment advantages in a tool interaction.}
Each invoke--assimilate pair passes through two distinct states: a \emph{transient} state $\tilde{s}_{k+1}$, containing the raw tool output immediately after invoke and the next \emph{persistent} state $s_{k+2}$, in which the assimilate segment has replaced raw output with its \texttt{<context>} block (formalized in Appendix~\ref{app:posmdp}). The critic $V_\phi$ evaluates at both, yielding separate advantages for query quality and extraction quality.
For a single tool call starting at boundary $k$, Equation~\ref{eq:segment_advantage} produces the following invoke--assimilate--commit triple:
\begin{align}
A(\text{invoke}) &= V_\phi(\tilde{s}_{k+1}) - V_\phi(s_k) & &\text{(query quality)}, \notag\\
A(\text{assimilate}) &= V_\phi(s_{k+2}) - V_\phi(\tilde{s}_{k+1}) & &\text{(extraction quality)}, \notag\\
A(\text{commit}) &= R - V_\phi(s_{k+2}) & &\text{(answer commitment)}.
\label{eq:per_segment_advantages}
\end{align}
All tokens within a segment share their segment's advantage, enabling invoke and assimilate to receive opposing credit within the same trajectory.

\subsection{The critic}
\label{sec:method:critic}

\paragraph{Architecture.}
A two-layer MLP value head attached to the shared LLM backbone produces $V_\phi(s) \in [0, 1]$ at each segment boundary. During critic-only warm-up, gradients update both the value head and backbone. The backbone learning rate is ten times smaller than the value-head learning rate (architecture details in Appendix~\ref{app:training}).

\paragraph{Training target.}
The critic is trained with Monte Carlo targets: MSE between $V_\phi(s)$ and the terminal reward $R$, averaged across boundary states. Tier labels determine the warm-up data mixture; every boundary in a rollout is trained against that rollout's final reward. MC targets are preferred because they are unbiased and their variance is bounded by the binary reward.

\paragraph{Warm-up.}
Critic warm-up before PPO has been demonstrated to help in prior RLHF work~\citep{zheng2023ppo_implementation} and is essential here: without it, early random critic gradients corrupt the shared backbone and the policy must unlearn them. The warm-up set is built from base-model rollouts labeled by final correctness, with no human annotation or external judge. Questions are classified as Tier~1 (beyond parametric competence) or Tier~2 (within it) via multi-rollout consistency and each is rolled out under forced-tool and no-tool prompts. This produces four outcome buckets that teach $V_\phi$ to separate what the model knows from what it does not (details in Appendix~\ref{app:warmup}). This warm-up is also applied to our Search-R1 PPO baseline; even with matched warm-up, segment-level credit retains a margin that grows with critic calibration (Appendix~\ref{app:warmup_ablation}). The cold-start ablation (no warm-up) trails our full method by ${\sim}$9 EM on HotpotQA at 7B (Appendix~\ref{app:warmup_ablation}). Figure~\ref{fig:calibration}(a) shows the resulting calibration curves: curated warm-up tracks the diagonal (ECE\,=\,0.043), random-data compresses the slope (ECE\,=\,0.137) and cold-start is near-flat (ECE\,=\,0.238).

\subsection{Training objective}
\label{sec:method:ppo}

With segments defined (\S\ref{sec:method:segments}), advantages derived (\S\ref{sec:method:advantage}) and the critic trained (\S\ref{sec:method:critic}), the training objective is standard PPO applied at the segment level. For each segment $k$, the PPO clipped surrogate objective is
\begin{equation}
\mathcal{L}^{\text{CLIP}}_k = \frac{1}{|g_k|} \sum_{t \in g_k} \min\!\Big(r_t(\theta)\, A(\text{seg}_k),\; \text{clip}\big(r_t(\theta),\, 1-\epsilon,\, 1+\epsilon\big)\, A(\text{seg}_k)\Big),
\label{eq:ppo_clip}
\end{equation}
where $g_k$ is the set of model-generated tokens in segment $k$ and $r_t(\theta) = \pi_\theta(a_t \mid c_t)/\pi_{\theta_{\text{old}}}(a_t \mid c_t)$ is the per-token importance ratio. Each segment receives its own advantage scalar, applied uniformly to all tokens within it. This enables invoke and assimilate to receive opposing credit when query quality and extraction quality disagree.

The critic is trained with the MC loss introduced in \S\ref{sec:method:critic}, formalized as $\mathcal{L}^{\text{critic}} = \frac{1}{N+1}\sum_{k=0}^{N}(V_\phi(s_k) - R)^2$. The core actor--critic objective combines the two: $\mathcal{L} = -\sum_{k=0}^{N}\mathcal{L}^{\text{CLIP}}_k + c_v \mathcal{L}^{\text{critic}}$, with the advantage treated as constant during policy-gradient computation. Optimization also includes a frozen-reference KL regularizer and an entropy bonus. Their coefficients are reported in Table~\ref{tab:hp}. After warm-up, the critic continues training alongside the policy using the outcomes of newly collected rollouts.

\section{Experiments}

We evaluated CARL on five benchmarks spanning two tool types, two Qwen2.5 model scales (3B and 7B) and a range of credit-assignment complexity. Our experiments answer three questions: does CARL outperform trajectory-level baselines (\S\ref{sec:exp:main}), does the critic actually learn the competence boundary we claim (\S\ref{sec:exp:critic}) and what bounds the remaining gains (\S\ref{sec:exp:errors})? Beyond accuracy and selectivity, we check whether tool training preserves existing capabilities using GSM8K and no-tool evaluations and whether the model remains robust when tools return unhelpful outputs (Appendix~\ref{app:capability}).

We additionally evaluate Llama 3.1 8B Instruct on the full benchmark pools, where CARL improves Avg$_3$ by 8.3 EM over Search-R1 GRPO while using 7.8\% fewer tokens (Appendix~\ref{app:llama-generalization}).

\subsection{Datasets}
\label{sec:exp:datasets}

We selected five benchmarks that jointly vary two axes: the typical number of tool calls per question and the fraction of questions the base model can answer from parametric knowledge alone. A \emph{hop} refers to one invoke--assimilate pair requiring an external tool call. We classify each question as \textbf{Tier~2} (within parametric competence) if the base model answers it correctly in at least one of five independent no-tool rollouts and \textbf{Tier~1} (beyond parametric competence) if all five rollouts produce incorrect answers. Table~\ref{tab:datasets} summarizes the five benchmarks.

\begin{table}[t]
\centering
\caption{Dataset characteristics. Tier~1 is the fraction of evaluation questions the base model answers incorrectly without tool access.}
\label{tab:datasets}
\small
\begin{tabular}{llcccc}
\toprule
Dataset & Tool & Hops & Tier~1 (3B) & Tier~1 (7B) & In training? \\
\midrule
GSM8K & Python & 1 & 8\% & 3\% & Yes \\
HotpotQA & BM25 search & 2 & 84\% & 76\% & Yes \\
2WikiMQA & BM25 search & 2 & 68\% & 57\% & Yes \\
FinQA & Python & 2--3 & 95\% & 91\% & No \\
MuSiQue & BM25 search & 2--4 & 96\% & 93\% & No \\
\bottomrule
\end{tabular}
\end{table}

GSM8K~\citep{cobbe2021gsm8k} is our pure selectivity test: almost entirely Tier~2, the question is whether the model learns to skip calling a tool. HotpotQA~\citep{yang2018hotpotqa} sits in the middle of the tool-necessity spectrum. Among the three search-based benchmarks, 2WikiMQA~\citep{ho2020wikimqa} has the highest Tier~2 rate (43\% at 7B), making it the strongest test of learning to skip searches the model does not need (Table~\ref{tab:datasets}). FinQA~\citep{chen2021finqa} is held out from training and tests generalization to heterogeneous tool routing (search followed by computation). MuSiQue~\citep{trivedi2022musique} is our hardest benchmark: compositional 3--4 hop questions, also held out.

\paragraph{Training data.} We train on the merged training splits of the three in-training datasets (HotpotQA, 2WikiMQA, GSM8K), totaling approximately 170K questions, matching the data scale of Search-R1~\citep{jin2025searchr1}. FinQA and MuSiQue are reserved entirely for out-of-distribution evaluation. For search benchmarks, we use Pyserini BM25 over a public prebuilt Wikipedia index with top-$k=3$; retrieval settings are provided in Appendix~\ref{app:reproducibility}.

\paragraph{Evaluation.} The main results use a fixed random sample of 500 questions per benchmark, drawn once and reused across methods, model scales and seeds. To limit the cost of full-pool tool-augmented inference, we restrict this validation to CARL, Search-R1 PPO and CoT+Tools (always) at both Qwen scales. We evaluate their existing checkpoints without retraining. The CARL--Search-R1 PPO Avg3 gains remain similar: 10.5 EM at 3B and 6.6 EM at 7B, compared with 10.6 and 6.7 on the sampled evaluation. Full-pool results are reported in Appendix~\ref{app:full_pool_evaluation}. Exact Match is computed after normalization. For multi-word entity answers, we also accept predictions whose normalized form contains the normalized gold as a substring.

\subsection{Methods}
\label{sec:exp:methods}

We compare four trained methods and one prompting baseline. Each of the three RL methods occupies a distinct point on the credit-assignment granularity spectrum: Search-R1 GRPO applies a single trajectory-level scalar to all tokens, Search-R1 PPO uses a token-level value function and CARL applies a segment-level value function at invoke/assimilate/commit boundaries. Reading the main table in order traces coarse $\to$ fine $\to$ boundary-aligned credit.

\paragraph{CoT + Tools.} A ReAct-style agent loop~\citep{yao2023react} over Qwen2.5-7B-Instruct with access to the same retrieval corpus as the trained methods (BM25, inline tables, Python); no fine-tuning. We run \emph{always-use} (tools on every question) and \emph{optional-use} (model discretion) variants as references for retrieval quality and prompting-only selectivity respectively.

\paragraph{SFT (rejection-sampled).} We supervised-fine-tune Qwen2.5-Instruct on its own correct (EM$=$1) tool-use trajectories, providing a supervised-learning baseline without RL credit assignment.

\paragraph{Search-R1 (GRPO and PPO).} We trained both variants in the open-source codebase~\citep{jin2025searchr1} on our training data with matched retrieval infrastructure, carefully tuning hyperparameters to give these baselines every advantage. For Search-R1 PPO, we additionally apply the same curated critic warm-up as CARL; we report results with and without warm-up in Appendix~\ref{app:warmup_ablation}, where even the strongest Search-R1 PPO configuration exceeds published numbers yet CARL retains a margin that grows with calibration.

\paragraph{CARL (ours).} Segment decomposition (\S\ref{sec:method:segments}) with segment-level PPO (\S\ref{sec:method:ppo}). Full hyperparameters are in Appendix~\ref{app:training}; retrieval setup, compute and reproducibility details are in Appendix~\ref{app:reproducibility}.

Methods using richer supervision (e.g. StepTool's GPT-4 judgments) operate under a different assumption and are not directly comparable; we discuss them in \S\ref{sec:rw} and Appendix~\ref{app:extended_rw}.

\subsection{Main results}
\label{sec:exp:main}

Table~\ref{tab:main_results} reports per-dataset accuracy and average tokens per episode across all methods and both scales. Search-R1 is a search-only framework and does not support non-search tools (e.g., calculator for GSM8K/FinQA); we therefore compare against Search-R1 PPO and GRPO on the three search-based datasets (HotpotQA, 2WikiMQA, MuSiQue). The table supports three claims.

We ablate CARL's credit assignment, invoke--assimilate separation, boundary placement, critic warm-up and terminal reward in Appendix~\ref{app:warmup_ablation}.

\definecolor{deltabg}{HTML}{E8F5E9}    
\definecolor{deltagreen}{HTML}{2E7D32}  
\definecolor{deltared}{HTML}{C62828}    
\definecolor{deltablue}{HTML}{1565C0}   
\newcommand{\pdt}[1]{{\scriptsize\color{deltagreen}\textbf{+#1}}}   
\newcommand{\ndt}[1]{{\scriptsize\color{deltared}\textbf{#1}}}      
\newcommand{\tdt}[1]{{\scriptsize\color{deltablue}\textbf{$\downarrow$#1}}}  
\begin{table}[t]
\centering
\caption{\textbf{Main results on fixed samples of 500 questions per benchmark.} Accuracy or EM ($\uparrow$) and generation length in tokens ($\downarrow$). $^\dagger$Training data; dashes indicate Search-R1 is not applicable. CARL and Search-R1 PPO report mean$\pm$std over 3 seeds; tokens are means. Prompting baselines, SFT and GRPO are single runs.}
\label{tab:main_results}
\setlength{\tabcolsep}{3pt}
\resizebox{\textwidth}{!}{%
\begin{tabular}{ll cc cc cc cc cc cc}
\toprule
& & \multicolumn{2}{c}{GSM8K} & \multicolumn{2}{c}{HotpotQA$^\dagger$} & \multicolumn{2}{c}{2Wiki$^\dagger$} & \multicolumn{2}{c}{FinQA} & \multicolumn{2}{c}{MuSiQue} & \multicolumn{2}{c}{Avg$_5$\,/\,Avg$_3$} \\
\cmidrule(lr){3-4} \cmidrule(lr){5-6} \cmidrule(lr){7-8} \cmidrule(lr){9-10} \cmidrule(lr){11-12} \cmidrule(lr){13-14}
Policy & Method & Acc & Tok & EM & Tok & EM & Tok & Acc & Tok & EM & Tok & Sc & Tok \\
\midrule
\multirow{6}{*}{\rotatebox{90}{\small 3B-Inst.}}
& CoT+Tools (always)        & 86.4 & 823  & 22.2 & 1943 & 19.4 & 2028 & 38.4 & 1962 & 4.6  & 2781 & 34.2\,/\,15.4 & 1907\,/\,2251 \\
& CoT+Tools (selective)     & 86.7 & 391  & 19.2 & 1685 & 21.2 & 1732 & 28.2 & 2096 & 5.3  & 2792 & 32.1\,/\,15.2 & 1739\,/\,2070 \\
& SFT (rej.\ sampled)       & 81.8 & 412  & 24.6 & 1818 & 25.2 & 1791 & 25.7 & 2014 & 4.8  & 2867 & 32.4\,/\,18.2 & 1780\,/\,2159 \\
& Search-R1 GRPO            & --   & --   & 34.4 & 1921 & 34.3 & 1960 & --   & --   & 8.6  & 2778 & --\,/\,25.8   & --\,/\,2220   \\
& Search-R1 PPO             & --   & --   & 30.2{\scriptsize$\pm$0.5} & 1809 & 32.1{\scriptsize$\pm$0.6} & 1861 & --   & --   & 12.1{\scriptsize$\pm$0.4} & 2755 & --\,/\,24.8   & --\,/\,2142   \\
& \textbf{CARL (ours)}      & \textbf{87.2}{\scriptsize$\pm$0.3} & \textbf{384}  & \textbf{40.4}{\scriptsize$\pm$0.2} & \textbf{1324} & \textbf{44.7}{\scriptsize$\pm$0.3} & \textbf{1394} & \textbf{46.6}{\scriptsize$\pm$0.4} & \textbf{1737} & \textbf{21.1}{\scriptsize$\pm$0.2} & \textbf{2546} & \textbf{48.0\,/\,35.4} & \textbf{1477\,/\,1755} \\
\rowcolor{deltabg}
& \textit{$\Delta$ vs.\ 2nd}  & \pdt{0.5} & \tdt{7} & \pdt{6.0} & \tdt{361} & \pdt{10.4} & \tdt{338} & \pdt{8.2} & \tdt{225} & \pdt{9.0} & \tdt{209} & \pdt{13.8\,/\,9.6} & \tdt{262\,/\,315} \\
\midrule
\multirow{6}{*}{\rotatebox{90}{\small 7B-Inst.}}
& CoT+Tools (always)        & 91.8 & 781  & 30.8 & 1862 & 26.6 & 1938 & 56.2 & 1623 & 11.4 & 2674 & 43.4\,/\,22.9 & 1776\,/\,2158 \\
& CoT+Tools (selective)     & 91.4 & 367  & 31.1 & 1621 & 27.1 & 1641 & 55.7 & 1785 & 9.8  & 2708 & 43.0\,/\,22.7 & 1624\,/\,1990 \\
& SFT (rej.\ sampled)       & 89.4 & 389  & 33.7 & 1712 & 32.6 & 1683 & 51.3 & 1864 & 8.9  & 2756 & 43.2\,/\,25.1 & 1681\,/\,2050 \\
& Search-R1 GRPO            & --   & --   & 39.1 & 1779 & 36.8 & 1827 & --   & --   & 18.5 & 2579 & --\,/\,31.5   & --\,/\,2062   \\
& Search-R1 PPO             & --   & --   & 42.8{\scriptsize$\pm$0.6} & 1837 & 41.7{\scriptsize$\pm$0.7} & 1775 & --   & --   & 22.1{\scriptsize$\pm$0.6} & 2682 & --\,/\,35.5   & --\,/\,2098   \\
& \textbf{CARL (ours)}      & \textbf{91.5}{\scriptsize$\pm$0.2} & \textbf{341}  & \textbf{47.9}{\scriptsize$\pm$0.3} & \textbf{1193} & \textbf{48.3}{\scriptsize$\pm$0.4} & \textbf{1139} & \textbf{59.3}{\scriptsize$\pm$0.4} & \textbf{1389} & \textbf{30.4}{\scriptsize$\pm$0.4} & \textbf{2511} & \textbf{55.5\,/\,42.2} & \textbf{1315\,/\,1614} \\
\rowcolor{deltabg}
& \textit{$\Delta$ vs.\ 2nd}  & \ndt{$-$0.3} & \tdt{26} & \pdt{5.1} & \tdt{428} & \pdt{6.6} & \tdt{502} & \pdt{3.1} & \tdt{234} & \pdt{8.3} & \tdt{68} & \pdt{12.1\,/\,6.7} & \tdt{309\,/\,376} \\
\bottomrule
\end{tabular}%
}
\par\vspace{4pt}
{\scriptsize\textit{Paired $\Delta$ (CARL$_i$ $-$ SR1-PPO$_i$, same seed) --- 7B: HotpotQA\,${+}5.1{\pm}0.3$, 2Wiki\,${+}6.6{\pm}0.5$, MuSiQue\,${+}8.3{\pm}0.4$; 3B: ${+}10.2{\pm}0.4$, ${+}12.6{\pm}0.4$, ${+}9.0{\pm}0.3$.}}
\end{table}

\paragraph{Testing the effect of credit assignment.}\footnote{Evaluation subsets vary by analysis. Table~\ref{tab:main_results} uses fixed samples of 500 questions per benchmark (2{,}500 questions total), with identical questions used across methods. Table~\ref{tab:selectivity} restricts to Tier~2 (parametrically solvable) questions from the three search benchmarks (${\sim}$370 questions) to isolate selectivity behavior. Intermediate evaluation during training uses the 2{,}500-question held-out evaluation pool. Figure~\ref{fig:calibration}(b) reports the four-benchmark average excluding GSM8K, which is calculator-only. To limit the cost of repeated tool-augmented inference, we use sampled evaluation rather than the complete benchmark pools to monitor training progress. Figure~\ref{fig:calibration}(c) plots Tier~2 tool-use rate across all five benchmarks.} At 7B, Search-R1 PPO exceeds Search-R1 GRPO by 4.0 average EM and CARL exceeds Search-R1 PPO by a further 6.7. These cross-method comparisons do not isolate the effect of credit assignment from other design differences.

CARL combines tool-aligned invoke, assimilate and commit boundaries, a context rewrite that replaces raw tool output, a warmed-up critic and segment-level credit. We examine credit computation, invoke--assimilate credit separation, boundary placement, critic warm-up and terminal reward through controlled variants. The interface and context-replacement mechanism remain in place; these experiments do not test removing assimilation itself.

\paragraph{Credit computation.} We compare segment-level TD advantages with token-level GAE and separately vary $\lambda$ at fixed segment boundaries (Appendix~\ref{app:lambda-ablation}).

\paragraph{Separate query and extraction credit.} We merge invoke and assimilate into one credited segment while retaining the context rewrite and raw-output discard (Appendix~\ref{app:merged-credit}).

\paragraph{Boundary placement.} We replace tool-event boundaries with count-matched newline or random-token boundaries (Appendix~\ref{app:boundary-placement}).

\paragraph{Critic warm-up.} We compare curated, random-data and cold-start conditions and vary the competence-label threshold, Tier~2/Tier~1 mixture and warm-up retrieval (Appendices~\ref{app:warmup_ablation} and \ref{app:warmup-sensitivity}).

\paragraph{Terminal reward.} We replace binary EM with graded F1 for both critic targets and PPO (Appendix~\ref{app:graded-reward}).

\paragraph{Evaluation-only competence check.} Separately, we re-evaluate the same held-out outputs under stricter competence definitions without retraining (Appendix~\ref{app:stricter-competence}).

\paragraph{Gains concentrate on multi-hop questions.} Per-dataset CARL gains over Search-R1 PPO at 7B are +5.1 on HotpotQA, +6.6 on 2WikiMQA and +8.3 on MuSiQue. The MuSiQue gain, on a 3--4 hop fully held-out benchmark, is the largest. The pattern repeats and amplifies at 3B: +10.2, +12.6 and +9.0 respectively, with the average 3B gain on these three search datasets (+10.6) running $1.6\times$ the corresponding 7B gain (+6.7) against Search-R1 PPO at both scales (Appendix~\ref{app:scale-mech}). Smaller models benefit disproportionately, consistent with the abstract's claim that knowing when to ask matters most when parametric memory is small (Appendix~\ref{app:scale-mech} decomposes this scaling asymmetry into three contributing mechanisms). SFT lifts in-domain HotpotQA and 2WikiMQA over the prompting ceiling (+2--6 EM at both scales) but does not transfer: it regresses on FinQA and is essentially neutral on MuSiQue, suggesting that imitation captures successful patterns but not selective tool use.

\paragraph{Tool-use efficiency improves substantially.} At 7B on the three PPO-comparable datasets, CARL averages 1{,}614 tokens per episode versus 2{,}098 for Search-R1 PPO, a 23\% reduction concentrated on the search tasks where calibrated $V(s_0)$ lets the model skip retrieval on Tier~2 questions\footnote{Tier~2 = answerable from parametric knowledge; Tier~1 = requires tool use to answer correctly. Definition: \S\ref{sec:exp:datasets}} (35--36\% fewer tokens on HotpotQA and 2WikiMQA). MuSiQue tokens remain high ($-$6\%) because its questions are inherently Tier~1 and require retrieval at every hop. Per-episode tool-call counts (Appendix~\ref{app:tool-calls}) confirm that the reduction is driven by selective invocation.

\subsection{How and why CARL works}
\label{sec:exp:critic}

\begin{figure*}[t]
  \centering
  \includegraphics[width=\textwidth]{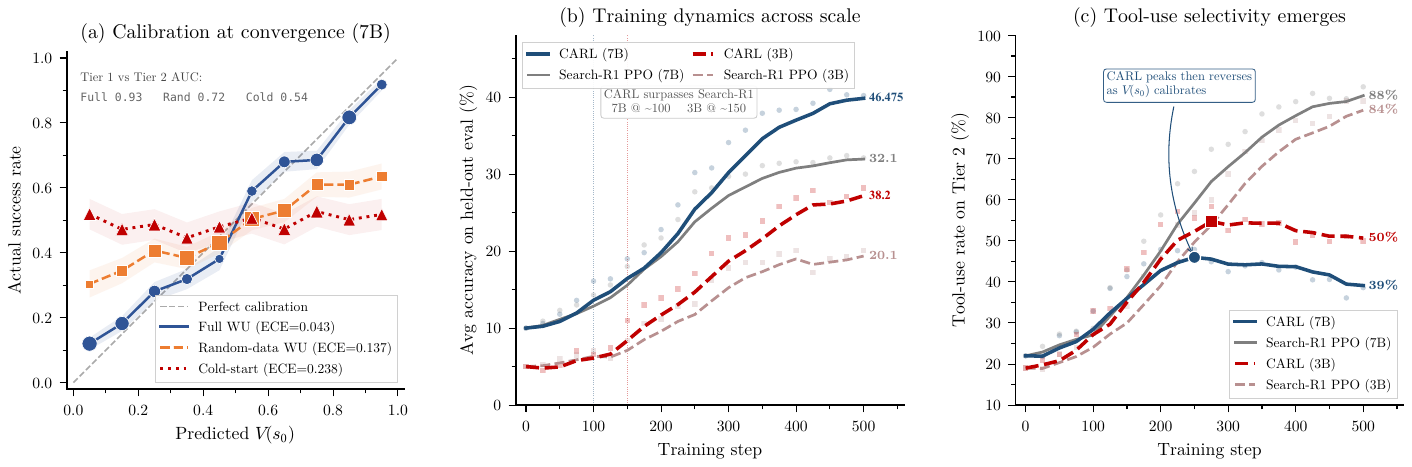}
  \caption{$V(s_0)$ calibration drives faster training and emergent selectivity. \textbf{(a)}~Calibration of $V(s_0)$ under three warm-up regimes (Appendix~\ref{app:warmup_ablation}). Full warm-up uses curated rollouts; random-data warm-up uses the same step budget without tier-based curation; cold start omits separate critic warm-up. \textbf{(b)}~Held-out accuracy across 500 PPO steps (4-dataset average from the 2{,}500-question evaluation pool). CARL surpasses Search-R1 PPO at step ${\sim}$100 (7B) / ${\sim}$150 (3B). \textbf{(c)}~Tier~2 tool-use rate. CARL peaks then settles (${\sim}$39\% at 7B); Search-R1 PPO climbs monotonically toward 84--88\%.}
  \label{fig:calibration}
\end{figure*}

We next show that the critic learns what the model knows and that this drives both faster training (Figure~\ref{fig:calibration}b) and selective tool use (Figure~\ref{fig:calibration}c, Table~\ref{tab:selectivity}).

\paragraph{A calibrated competence boundary.} $V(s_0)$ classifies Tier~1 from Tier~2 questions with AUC 0.93 at 7B and 0.85 at 3B (Figure~\ref{fig:vs0_distribution}; calibration shape in Figure~\ref{fig:calibration}(a)). Calibration matters as well as ranking: random-data warm-up reaches AUC 0.72 but ECE 0.137, producing advantages that carry the right sign more often than the right magnitude. The critic also signs intermediate steps correctly: $V$ rises after successful extraction 79.4\% of the time and drops when assimilation loses a surfaced answer (72.1\%), a within-trajectory sign divergence that trajectory-level methods cannot produce (per-segment sign statistics in Appendix~\ref{app:sign-behavior}, Table~\ref{tab:sign-behavior}).

\begin{figure*}[t]
  \centering
  \includegraphics[width=\textwidth]{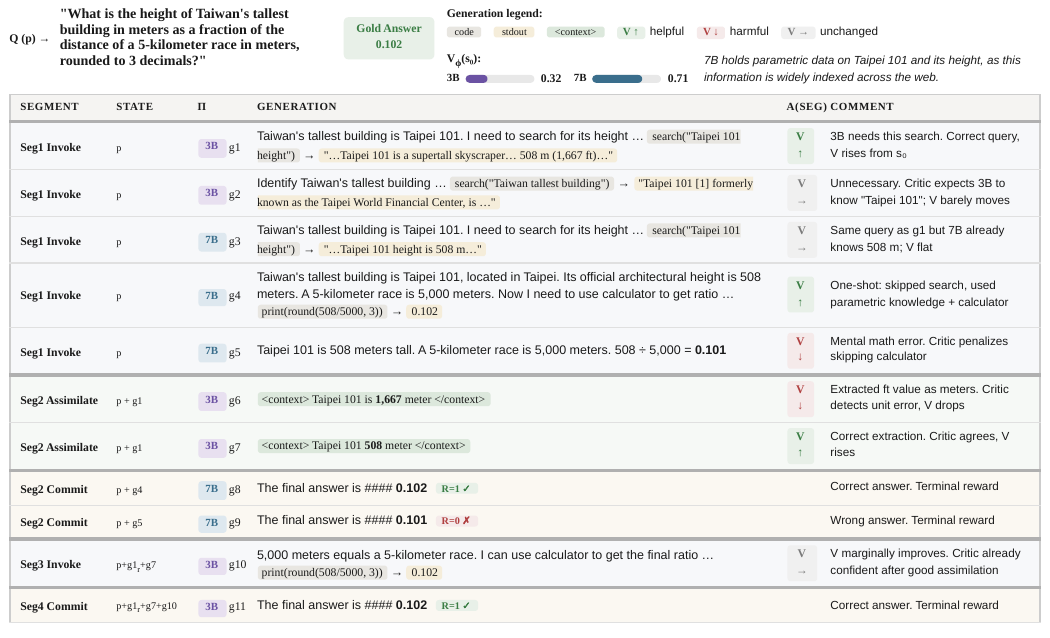}
  \caption{\textbf{Qualitative analysis of the CARL critic on a compositional query.} The same prompt elicits distinct critic behaviour at each scale ($V(s_0)$: 3B\,=\,0.32, 7B\,=\,0.71). \textbf{(i)}~A search raises $V$ at 3B but not 7B, which already knows the answer. \textbf{(ii)}~Faithful extraction raises $V$; a unit-error extraction lowers it, showing opposing signs within one trajectory. \textbf{(iii)}~At 7B, calculator use earns $V{\uparrow}$ while mental math earns $V{\downarrow}$. All advantages arise from binary reward alone.}
  \label{fig:trajectories}
\end{figure*}

\paragraph{Calibration drives faster training.} Once explained variance crosses 0.60 (around step ${\sim}$100 at 7B, ${\sim}$150 at 3B), CARL pulls ahead of Search-R1 PPO (Figure~\ref{fig:calibration}(b)) and the gap widens monotonically. Before this point, segment-level advantages are noisy enough that token-level credit performs comparably; after it, they reliably point in the right direction. Explained variance rises from 0.52 after warm-up to 0.79 at convergence (Appendix~\ref{app:ev-trajectory}, Table~\ref{tab:ev-trajectory}). Warm-up amplifies the gap monotonically: CARL leads Search-R1 PPO at every warm-up level (cold, random, curated) and the margin grows from $+1.7$ to $+5.1$ EM on HotpotQA at 7B as calibration improves (Appendix~\ref{app:warmup_ablation}).

\begin{table}[t]
\centering
\caption{Tool-use selectivity at convergence (Qwen2.5-7B), evaluated over Tier 2 questions from HotpotQA, 2WikiMQA and MuSiQue ($\sim$370 questions total). These are the three datasets where Search-R1 was trained in its public release, so all comparisons are apples-to-apples. Tool rate is the percentage of questions on which at least one tool is invoked. Tier 1 tool rate confirms CARL still invokes tools when needed. Tier 2 tool rate measures unnecessary tool calls. Tier 2 EM measures accuracy on these questions, regardless of whether the method invoked tools.}
\label{tab:selectivity}
\small
\begin{tabular}{l ccc}
\toprule
Method & Tier 1 tool rate & Tier 2 tool rate & Tier 2 EM \\
\midrule
CoT+Tools (always) & 100.0\% & 100.0\% & 72.4 \\
Search-R1 GRPO     &  96.2\% &  87.5\% & 75.1 \\
Search-R1 PPO      &  95.4\% &  82.6\% & 77.2 \\
\textbf{CARL (ours)} &  \textbf{97.8\%} & \textbf{38.6\%} & \textbf{87.0} \\
\bottomrule
\end{tabular}
\end{table}

\paragraph{Calibration drives emergent selectivity.} Because $V(s_0)$ is high on Tier~2 questions, the invoke advantage on those questions is small by construction and the policy receives a stronger gradient from answering directly. This is the mechanism producing the gap in Figure~\ref{fig:calibration}(c) and the convergence numbers in Table~\ref{tab:selectivity}: at 7B, CARL invokes tools on 38.6\% of Tier~2 questions versus 82.6\% for Search-R1 PPO, despite using no explicit uncertainty heuristic, indicating that selectivity emerges from $V(s_0)$, a learned competence estimate. Despite calling tools far less often, CARL achieves 87.0 Tier~2 EM compared to 77.2 for Search-R1 PPO and 75.1 for GRPO, achieving fewer tool calls and higher accuracy on the same questions. CARL also reduces hallucination on Tier~1 questions, where $V(s_0)$ consistently routes to tool use rather than fabrication (Appendix~\ref{app:hallucination}).

The 3B selectivity peak is higher and later because the less sharp $V(s_0)$ (AUC 0.85) provides less counter-pressure; 3B gains exceed 7B at every hop count (Appendix~\ref{app:hops}, \ref{app:scale-mech}).

\subsection{What bounds CARL's gains}
\label{sec:exp:errors}

Manual analysis on MuSiQue (Appendix~\ref{app:errors}) finds that 48\% of CARL's remaining errors trace to irrelevant BM25 retrieval, with another 14\% to query formulation; combined, retrieval-side failures account for 62\% of remaining errors, with the dominant category (irrelevant retrieval) stable across hop counts (47--50\%). Replacing BM25 with a commercial search API (Appendix~\ref{app:retrieval}) lifts CARL's MuSiQue accuracy from 30.4 to 45.0 EM, bringing a 7B open-weight model into the range typically reported for substantially larger models on MuSiQue. These results suggest retrieval quality, not the credit-assignment method, sets the ceiling on MuSiQue.

\section{Limitations}
\label{sec:limitations}

\textbf{Warm-up cost.} Segment-level credit requires a reasonably calibrated critic before PPO. Our warm-up procedure (\S\ref{sec:method:critic}) mitigates the cold-start problem, but an initial training phase, during which the critic learns from base-model rollouts without policy updates, remains an inherent overhead. Moreover, the warm-up ablation (Appendix~\ref{app:warmup_ablation}) shows the benefit of curated warm-up scales with model size: at 3B even random-data warm-up suffices for CARL to lead Search-R1 PPO, while at 7B curated warm-up is needed to realize the full $+5.1$ EM margin (Appendix~\ref{app:warmup_ablation} Table~\ref{tab:warmup_ablation}); CARL still leads at every warm-up level, but the margin shrinks from $+5.1$ to $+1.7$ EM as warm-up degrades.

\textbf{Segment granularity.} All tokens within a segment share one advantage value, so intra-segment variation (e.g., a partly useful extraction) is not captured; credit is uniform within each segment even when individual tokens contribute unequally.

\textbf{Computational overhead.} Each tool call adds two segments and corresponding critic forward passes. Current serving stacks (vLLM, TGI, SGLang) are optimized for uninterrupted decode-to-EOS and do not natively support mid-generation pause, K/V cache surgery, or synchronous tool calls, so segment boundaries currently require per-segment round-trips and partial re-prefill. This is an engineering gap in the serving layer rather than a methodological one and we expect it to narrow as inference engines add first-class tool-call primitives.

\section{Conclusion}

Tool-use trajectories have structurally observable boundaries that pure reasoning trajectories lack. CARL exploits this structure: a critic trained on the model's own rollouts turns binary outcome reward into per-segment advantages grounded in the SMDP Bellman equation, giving each tool interaction its own credit without external judges or step-level annotations. The result is a model that learns not just \emph{how} to use tools but \emph{when}, issuing fewer unnecessary calls, formulating better queries and extracting more faithfully, with gains concentrating on multi-hop tasks where credit assignment is hardest.

Several directions follow naturally. The invoke/assimilate/commit decomposition applies unchanged to richer tool ecosystems (code execution, API calls, multi-agent coordination) where trajectory-level credit becomes even less informative. The competence boundary encoded in $V(s_0)$ could serve as a routing signal at inference time, dispatching to specialized tools or larger models only when predicted success probability is low. More broadly, our finding that smaller models benefit disproportionately suggests that segment-level credit assignment is a practical alternative to scaling parametric memory for tail-distribution knowledge, particularly relevant for resource-constrained settings such as on-device assistants and local coding agents where smaller models with strong tool-use behavior offer a viable alternative to API-bound large models.

\clearpage
\begin{ack}
This work was supported by Microsoft. The authors declare no competing interests.
\end{ack}

\newpage
\appendix

\section{SMDP derivation details}
\label{app:smdp_derivation}

\paragraph{From the SMDP Bellman equation.}
By Theorem~1 of \citet{sutton1999options}, our setting is an SMDP over segment-level decision points, with per-segment TD error
\begin{equation}
\delta^{\text{SMDP}}_k = r_k + \gamma^{\tau_k}\, V(s_{k+1}) - V(s_k),
\label{eq:smdp_td}
\end{equation}
where $r_k$ is the cumulative reward during segment $k$ and $\tau_k$ is its length in tokens. Two properties of our setting simplify this expression. First, the task is episodic with finite horizon (at most 15 segments), so we set $\gamma = 1$ and $\gamma^{\tau_k} = 1$ for every segment. Second, reward is terminal-only: $r_k = 0$ for $k < N$ and $r_N = R$. Substituting into Equation~\ref{eq:smdp_td} gives the per-segment advantage in Equation~\ref{eq:segment_advantage}.

\paragraph{Total credit is preserved.}
Summing the per-segment advantages along a trajectory, the intermediate $V_\phi$ terms cancel pairwise, leaving only the endpoints:
\begin{equation}
\sum_{k=0}^{N} A(\text{seg}_k) \;=\; \big[V_\phi(s_1) - V_\phi(s_0)\big] + \big[V_\phi(s_2) - V_\phi(s_1)\big] + \cdots + \big[R - V_\phi(s_N)\big] \;=\; R - V_\phi(s_0).
\label{eq:telescope}
\end{equation}
This is the same total credit that trajectory-level PPO would assign from the starting state. Segment-level credit does not add or remove credit; it redistributes the same total across segments, crediting or penalizing each according to how much it changed the critic's estimate. The comparison between CARL and trajectory-level baselines is therefore a comparison of \emph{how} credit is distributed, not \emph{how much}.\footnote{The intermediate advantage $V_\phi(s_{k+1}) - V_\phi(s_k)$ has the potential-based form $\Phi(s') - \Phi(s)$ of \citet{ng1999policyinvariance}, with the critic $V_\phi$ playing the role of the potential $\Phi$. Because the potential is learned jointly with the policy rather than fixed in advance, the full policy-invariance guarantee holds only as $V_\phi$ stabilizes; but the form motivates our construction and matches the shape of process rewards used in prior work (\S\ref{sec:rw:credit}).}

\paragraph{Why $\lambda = 0$.} One could apply GAE at the segment level: $A^{\text{GAE}}_k = \delta_k + \lambda \delta_{k+1} + \lambda^2 \delta_{k+2} + \cdots$. We use $\lambda = 0$ (the one-step TD error) to avoid mixing later segments into the current segment's credit.

Segment-level credit assignment relies on isolating the contribution of each segment; multi-step lookahead reintroduces the temporal smearing that segment-level credit is designed to avoid. A bad invoke followed by a strong recovery would see its negative advantage diluted by future positive residuals.

\paragraph{The $\lambda \to 1$ limit collapses to trajectory-level credit.} Applying GAE \citep{schulman2015high} at the segment level with discount $\gamma = 1$ gives
\begin{equation}
A^{\text{GAE}}(\text{seg}_k) \;=\; \sum_{l=0}^{N-k} \lambda^l \, \delta_{k+l}, \qquad \delta_j = r_j + V_\phi(s_{j+1}) - V_\phi(s_j).
\label{eq:gae-segment}
\end{equation}
Under terminal-only reward ($r_j = 0$ for $j < N$, $r_N = R$, with $V_\phi(s_{N+1}) \equiv 0$), this gives $\delta_j = V_\phi(s_{j+1}) - V_\phi(s_j)$ for $j < N$ and $\delta_N = R - V_\phi(s_N)$. At $\lambda = 1$, Eq.~\ref{eq:gae-segment} becomes a telescoping sum:
\begin{align}
A^{\text{GAE}}(\text{seg}_k) \big|_{\lambda=1}
\;&=\; \sum_{l=0}^{N-k} \delta_{k+l} \nonumber \\
\;&=\; \underbrace{\bigl[V_\phi(s_{k+1}) - V_\phi(s_k)\bigr]}_{\delta_k} + \underbrace{\bigl[V_\phi(s_{k+2}) - V_\phi(s_{k+1})\bigr]}_{\delta_{k+1}} + \cdots + \underbrace{\bigl[R - V_\phi(s_N)\bigr]}_{\delta_N} \nonumber \\
\;&=\; R - V_\phi(s_k).
\label{eq:lambda-one-collapse}
\end{align}
Every intermediate $V_\phi$ term cancels pairwise; only the endpoints $-V_\phi(s_k)$ and $R$ survive.

With $\lambda = 1$ and $\gamma = 1$, summing TD errors through the terminal outcome gives each segment the advantage $R-V_\phi(s_k)$, assuming no intermediate rewards and zero value after termination. If the terminal reward is binary and critic values lie in $[0,1]$, this imposes the following sign restriction on the raw advantages:
\begin{itemize}[leftmargin=*,itemsep=2pt]
\item In a successful trajectory ($R = 1$), $A(\text{seg}_k) = 1 - V_\phi(s_k) \geq 0$ for every $k$. Every segment receives non-negative credit.
\item In a failed trajectory ($R = 0$), $A(\text{seg}_k) = -V_\phi(s_k) \leq 0$ for every $k$. Every segment receives non-positive credit.
\end{itemize}
Under these conditions, segments within the same trajectory cannot receive opposite-sign advantages, although their magnitudes can differ. A useful retrieval cannot receive positive credit in a failed trajectory and a faulty extraction cannot receive negative credit in a successful one. This restriction does not apply to CARL's default $\lambda = 0$, where intermediate advantages measure the value change across each segment. It also need not hold for graded terminal rewards.

Allowing opposing segment advantages under binary rewards therefore requires $\lambda < 1$, not necessarily exactly zero. The three-seed 3B sweep in Appendix~\ref{app:lambda-ablation} shows comparable Avg$_5$ at $\lambda = 0$ and $0.1$ (48.0 and 48.1), falling to 40.9 at $\lambda = 1$.

One-step TD has lower variance but is more exposed to critic approximation error, making critic calibration important. CARL uses MC terminal-outcome targets to warm up the critic before policy training (Appendix~\ref{app:warmup}).

\section{Partially-observed SMDP formalism}
\label{app:posmdp}

In CARL's default context-management mode (code replacement + assimilation), the raw tool output $o_k^{\text{raw}}$ is observed by the assimilate segment but removed from context before the next segment begins. This section formalizes the resulting partially-observed SMDP and shows that segment-level advantages remain well-defined.

\paragraph{Transient and persistent states.}
Let $s_k$ denote the \emph{persistent} state at segment boundary~$k$, i.e.\ the full context visible to all subsequent segments. After an invoke segment, the rollout machinery appends the code output, producing a \emph{transient} state
\begin{equation}
\tilde{s}_{k+1} = s_k \oplus t_k \oplus o_k^{\text{raw}},
\end{equation}
where $t_k$ is the model-generated text (planning + code) and $o_k^{\text{raw}}$ is the raw stdout. The assimilate segment conditions on $\tilde{s}_{k+1}$ and produces an extraction $e_{k+1}$. The rollout loop then replaces $o_k^{\text{raw}}$ with $e_{k+1}$, yielding the next persistent state:
\begin{equation}
s_{k+2} = s_k \oplus t_k \oplus e_{k+1}.
\end{equation}

\paragraph{Observation model.}
The mapping $\tilde{s}_{k+1} \mapsto s_{k+2}$ is \emph{not} a fixed function of the environment; it depends on the policy's assimilation output $e_{k+1}$. This means the transition from boundary~$k$ to boundary~$k{+}2$ is partially observed: the invoke segment's raw output is visible only to the immediately following assimilate segment, not to any later segment.

\paragraph{Segment-level advantages remain well-defined.}
The critic $V_\phi$ is evaluated at persistent-state boundaries $s_0, \tilde{s}_1, s_2, \tilde{s}_3, s_4, \ldots$ (alternating transient and persistent for tool-call trajectories). Each advantage $A(\text{seg}_k) = V_\phi(s_{k+1}) - V_\phi(s_k)$ depends only on the two boundary states of that segment, both of which are observed by the critic at evaluation time. The partial observability affects what information future segments can condition on, but not the well-definedness of the advantage at the segment where $V_\phi$ is computed. The telescoping property (Equation~\ref{eq:telescope}) holds identically: intermediate $V_\phi$ terms cancel regardless of whether boundary states are transient or persistent.

\paragraph{Why gradient masking is unnecessary.}
In Search-R1 and ToRL, environment-injected tokens (tool output) appear in the generation context and must be masked from the policy gradient to avoid reinforcing tokens the model did not produce. In CARL, raw tool output $o_k^{\text{raw}}$ is replaced by $e_{k+1}$ before the next generation begins, so the only tokens in the policy's generation target are model-produced tokens. Each segment's loss (Equation~\ref{eq:ppo_clip}) sums over $g_k$, the model-generated tokens in that segment; environment tokens never enter $g_k$. Gradient masking is therefore unnecessary by construction.

\section{Warm-up implementation details}
\label{app:warmup}

\paragraph{Dataset sizes per bucket.}
The warm-up dataset is constructed from base-model rollouts on the training split. Table~\ref{tab:warmup_counts} reports the number of trajectories per bucket.

\begin{table}[H]
\centering
\small
\caption{Warm-up dataset composition (number of trajectories). Both scales use 32K trajectories total, but the Tier 1/Tier 2 split differs because it is determined by each base model's competence: the 7B model answers more questions correctly without tools (45\% Tier~2), while the 3B model fails more often (35\% Tier~2). Within each tier, trajectories are split equally between no-tool and forced-tool rollouts.}
\label{tab:warmup_counts}
\begin{tabular}{lrr}
\toprule
Bucket & 7B & 3B \\
\midrule
Tier~2, no-tool (anchors $V(s_0) \approx 1$) & 7{,}200 & 5{,}600 \\
Tier~2, forced-tool (unnecessary tool risk) & 7{,}200 & 5{,}600 \\
Tier~1, no-tool (anchors $V(s_0) \approx 0$) & 8{,}800 & 10{,}400 \\
Tier~1, forced-tool (teaches assimilation) & 8{,}800 & 10{,}400 \\
\midrule
Total & 32{,}000 & 32{,}000 \\
\bottomrule
\end{tabular}
\end{table}

The Tier~2/Tier~1 ratio reflects each base model's competence boundary: 45/55 at 7B and 35/65 at 3B (averaging $\sim$40/60 across scales), with roughly 70\% BM25 retrieval and 30\% high-quality search API retrieval within each bucket.

\paragraph{Curation principles.}
The warm-up set is built from base-model rollouts, labeled by final correctness. No human annotation and no external judge are used.
\begin{enumerate}[leftmargin=*,itemsep=1pt]
\item \textbf{Tier 1 / Tier 2 labeling via multi-rollout consistency.} A question is Tier~2 (within parametric competence) if the base model answers it correctly in at least one of five no-tool rollouts and Tier~1 (beyond parametric competence) if all five rollouts produce incorrect answers. This labeling directly measures the model's competence boundary: Tier~2 questions are those the model can sometimes answer from its own knowledge; Tier~1 questions are those where it consistently cannot. Because the 7B model answers more questions correctly, its warm-up set is 45\% Tier~2; the 3B set is 35\% Tier~2. Both force $V_\phi$ to discriminate along the competence axis.
\item \textbf{Controlled tool-use rollouts.} Each question is also rolled out under two system prompts: (a) forced tool use, (b) no tools allowed. Combined with the Tier labels, this produces four outcome buckets per question, each teaching a different signal: Tier~2 no-tool (anchors $V(s_0) \approx 1$), Tier~2 forced-tool (teaches that unnecessary tools add risk), Tier~1 no-tool (anchors $V(s_0) \approx 0$), Tier~1 forced-tool (teaches assimilation).
\item \textbf{Mixed retrieval quality.} Roughly 70\% BM25 (matching PPO training) and 30\% rollouts from a high-quality internal search API give $V_\phi$ contrast between helpful and less-helpful tool outputs without changing the inference-time retriever.
\item \textbf{Topic diversity via embedding clustering.} Questions are embedded with a sentence-transformer (all-MiniLM-L6-v2) and clustered with $k$-means; warm-up samples are drawn uniformly across clusters to prevent $V_\phi$ from learning surface topical cues.
\item \textbf{Multi-hop exposure.} Multi-hop rollouts from 2WikiMQA are included so that $V_\phi$ has seen compound-state boundaries before PPO.
\end{enumerate}

\paragraph{Verification gate.}
Before starting PPO, three checks must pass on a held-out subset.

\textit{Minimum thresholds (gate):} (i) $V(s_0)$ separates Tier~1 from Tier~2 questions (AUC $\geq 0.70$), (ii) $V_\phi$ shows correct sign behavior after retrieval on forced-tool rollouts ($\geq 60\%$ of cases) and (iii) the critic achieves sufficient explained variance (EV $\geq 0.45$).

\textit{Target thresholds (well-warmed-up critic):} A well-warmed-up critic typically clears the gate by a comfortable margin (AUC $\geq 0.80$, sign accuracy $\geq 70\%$, EV $\geq 0.55$). Random-data warm-up alone reaches AUC 0.72, only just above the 0.70 gate, so the gate is necessary but not sufficient: calibration quality (ECE) matters beyond ranking ability.

\paragraph{What the ablations show.}
The three warm-up variants produce qualitatively different critic shapes. Full warm-up yields a near-diagonal calibration curve (ECE = 0.043), random-data warm-up produces a compressed monotone curve (ECE = 0.137) and cold-start gives a flat near-chance line (ECE = 0.238).
Figure~\ref{fig:calibration}(a) plots predicted success probability $V(s_0)$ against actual success rate on held-out questions. Full warm-up tracks the diagonal closely; random-data warm-up ranks correctly but pulls predictions toward the mean (slope $\approx$ 0.4), concentrating calibration error at the extremes; cold-start predictions are nearly flat, consistent with AUC $\approx$ 0.54 (near chance). Figure~\ref{fig:calibration}(b) plots held-out evaluation accuracy throughout training: curated warm-up pulls ahead of Search-R1 PPO within the first hundred steps, random-data warm-up lags and cold-start plateaus below both.

\paragraph{System prompt templates.}
Four system prompts are used during warm-up data collection:

\begin{enumerate}[leftmargin=*,itemsep=2pt]
\item \textbf{No-tool direct answer:} ``Answer the following question directly. Do not use any tools or code. Provide your answer inside \textbackslash boxed\{\}.''
\item \textbf{Forced tool use:} ``You must use a Python code block to help answer this question. Use \texttt{search(query)} for retrieval or write arithmetic code. After seeing tool output, write a \texttt{<context>} block extracting the relevant information, then provide your answer inside \textbackslash boxed\{\}.''
\item \textbf{Optional tool use (Tier~2):} Same as forced tool use but with ``You may optionally use'' replacing ``You must use.''
\item \textbf{Multi-hop forced tool use:} Same as (2) but with an additional instruction: ``This question may require multiple search steps. You may call tools more than once.''
\end{enumerate}

\paragraph{Embedding clustering.}
Questions are embedded with \texttt{all-MiniLM-L6-v2} (384-dimensional embeddings). We apply $k$-means with $k = 50$ clusters for HotpotQA and 2WikiMQA, $k = 30$ for GSM8K (smaller and more homogeneous). Warm-up samples are drawn uniformly across clusters within each bucket to prevent the critic from learning surface topical cues. Clustering is performed once on the full training split; the same cluster assignments are reused for both scales.

\paragraph{Training schedule.}
Warm-up training continues until the verification gate passes on a held-out subset: AUC $\geq 0.70$, sign accuracy $\geq 0.60$ and explained variance $\geq 0.45$. We checkpoint every 25 steps and use the first checkpoint at or after step 1{,}800 that clears the gate. In practice this requires 1{,}800--2{,}400 critic-only steps depending on scale and all 6 trained models (2 scales $\times$ 3 seeds) cleared the gate within this range. Each step processes a batch of 256 (trajectory, boundary) state-target pairs sampled uniformly across the four warm-up buckets, with MSE loss against Monte-Carlo terminal returns. This corresponds to roughly 5--6 epochs over the boundary-state pool ($\sim$96K boundaries from 32K trajectories at $\sim$3 boundaries each). The optimizer is AdamW ($\beta_1{=}0.9$, $\beta_2{=}0.999$) with weight decay $0.0$ for the value head and $0.01$ for the backbone. We set value-head weight decay to zero because L2 regularization on a sigmoid output shrinks predictions toward 0.5, undermining calibration. Learning rates are 5e-6 for the value head and 5e-7 for the backbone ($10\times$ smaller, to preserve policy representations), with linear warmup over the first 100 steps and constant thereafter. Gradient clipping is set at $\|\nabla\|_2 = 1.0$. No PPO policy-gradient updates are applied during warm-up; critic gradients update the value head and shared backbone.

\begin{table}[H]
\centering
\small
\caption{Warm-up defaults. Bucket counts are in Table~\ref{tab:warmup_counts}; optimizer settings are in Table~\ref{tab:hp}.}
\label{tab:warmup-defaults}
\begin{tabular}{p{0.24\textwidth}p{0.68\textwidth}}
\toprule
Setting & Default \\
\midrule
Competence label & Tier~2: at least one correct answer in five no-tool rollouts; Tier~1: all five incorrect. \\
Training set & 32K trajectories; equal no-tool and forced-tool shares within each tier. \\
Tier~2/Tier~1 mixture & 35/65 at 3B; 45/55 at 7B. \\
Retrieval mixture & Approximately 70\% BM25 and 30\% search API. \\
Embedding and clusters & \texttt{all-MiniLM-L6-v2}; $k=50$ for HotpotQA/2WikiMQA and $k=30$ for GSM8K. \\
Critic fitting & 1{,}800--2{,}400 steps; batch size 256; MSE against MC terminal outcomes. \\
Learning rates & $5\times10^{-6}$ for the value head; $5\times10^{-7}$ for the backbone. \\
Minimum gate & AUC $\geq0.70$, sign accuracy $\geq60\%$ and EV $\geq0.45$. \\
\bottomrule
\end{tabular}
\end{table}

\begin{table}[H]
\centering
\small
\caption{Approximate warm-up cost per model size in H100-equivalent GPU-hours. Collection generations and critic fitting are accounted for separately.}
\label{tab:warmup-cost}
\begin{tabular}{lrrr}
\toprule
Stage & Generations & 3B hours & 7B hours \\
\midrule
Embedding and clustering & 0 & $<1$ & $<1$ \\
Tier labeling (16K questions $\times$ 5) & 80{,}000 & ${\sim}5$ & ${\sim}12$ \\
Forced-tool rollouts (one per question) & 16{,}000 & ${\sim}130$ & ${\sim}300$ \\
Critic fitting (${\sim}$2{,}000 steps) & 0 & ${\sim}50$ & ${\sim}110$ \\
\midrule
Total & 96{,}000 & ${\sim}185$ & ${\sim}420$ \\
\bottomrule
\end{tabular}
\end{table}

\paragraph{Example warm-up rollouts.}
Representative trajectories from each bucket, showing the critic's $V_\phi$ at each segment boundary, are provided in the supplementary materials.

\section{Training hyperparameters and compute}
\label{app:training}

Table~\ref{tab:hp} reports the full training configuration. Search-R1 PPO is trained with the verl-based public codebase released by~\citet{jin2025searchr1}. Shared algorithmic values are matched across both implementations; implementation-specific names are given in parentheses for reproducibility.

\paragraph{Step accounting and data exposure.}
We use the term \emph{PPO step} (or simply \emph{step}) to mean one outer iteration: \textsc{Rollout} (sample 256 prompts and generate $n{=}5$ trajectories each, for 1{,}280 trajectories per step) $\to$ \textsc{Reward/Advantage} (EM check, critic evaluation at boundaries, per-segment advantages) $\to$ \textsc{Learn} (4 PPO inner epochs over the 1{,}280 trajectories; with mini-batch 64, this is 80 gradient updates per step). Across 500 steps, training therefore processes 128K unique prompts (0.75 epoch over the 170K training pool), generates 640K trajectories total and performs 40K gradient updates. The 0.75-epoch coverage is in line with published RL-LLM practice~\citep{jin2025searchr1,yu2025dapo} where prompt-level sample efficiency is governed by gradient noise rather than data exhaustion.

\begin{table}[H]
\centering
\scriptsize
\caption{Training hyperparameters. Search-R1 PPO baseline reproduced from its public verl codebase. Where a name differs across implementations, both are shown.}
\label{tab:hp}
\setlength{\tabcolsep}{1.5pt}
\begin{tabular}{lll}
\toprule
Hyperparameter & CARL & Search-R1 PPO (verl) \\
\midrule
\multicolumn{3}{l}{\textit{PPO algorithm}} \\
PPO clip ratio (\texttt{eps\_clip})           & 0.2                       & 0.2 \\
GAE $\lambda$ (\texttt{lambd})                & 0.0 (one-step TD, \S\ref{sec:method:advantage}) & 0.95 \\
Discount $\gamma$                              & 1.0                       & 1.0 \\
Value loss coef (\texttt{vf\_coef})           & 0.5                       & 0.5 \\
Entropy coef                                   & 0.001                     & 0.001 \\
KL penalty coef (\texttt{init\_kl\_coef})     & 0.001                     & 0.001 \\
KL estimator                                   & k3 (\texttt{use\_kl\_estimator\_k3}) & low\_var\_kl \\
KL as loss term                                & \texttt{use\_kl\_loss=true} & \texttt{use\_kl\_loss=true} \\
\midrule
\multicolumn{3}{l}{\textit{Batch sizes}} \\
Rollout batch (\texttt{rollout\_batch\_size}, prompts/step) & 256 & 256 (\texttt{data.train\_batch\_size}) \\
Micro-rollout batch per GPU (\texttt{micro\_rollout\_batch\_size}) & 8 & --- (dynamic) \\
Train batch (\texttt{train\_batch\_size}, PPO mini-batch) & 64 & 64 (\texttt{ppo\_mini\_batch\_size}) \\
Micro-train batch per GPU (\texttt{micro\_train\_batch\_size}) & 2 & 2 (\texttt{ppo\_micro\_batch\_size\_per\_gpu}) \\
Rollouts per prompt (\texttt{n\_samples\_per\_prompt}) & 5 & 5 (\texttt{rollout.n}) \\
Trajectories per RL step                       & 1{,}280                  & 1{,}280 \\
PPO inner epochs (\texttt{max\_epochs})       & 4                         & 4 (\texttt{ppo\_epochs}) \\
\midrule
\multicolumn{3}{l}{\textit{Sequence lengths}} \\
Prompt max len (\texttt{prompt\_max\_len})    & 2048                     & 2048 (\texttt{max\_prompt\_length}) \\
Generate max len (\texttt{generate\_max\_len}) & 2048 per decision segment & 2048 (\texttt{max\_response\_length}) \\
Max segments per episode (CARL only)           & 15 including assimilation & --- \\
$T_{\text{assimilate}}$                        & 256 per assimilation segment & --- \\
\midrule
\multicolumn{3}{l}{\textit{Generation (rollout sampling)}} \\
Temperature                                    & 1.0                       & 1.0 \\
Top-$p$                                        & 1.0                       & 1.0 \\
Top-$k$                                        & $-1$ (off)                & $-1$ \\
Rollout backend                                & vLLM (\texttt{--vllm\_num\_engines 4}) & vLLM (\texttt{rollout.name=vllm}) \\
\midrule
\multicolumn{3}{l}{\textit{Learning rates (constant after 100-step linear warm-up)}} \\
Actor LR (\texttt{actor\_learning\_rate})     & \shortstack[l]{1e-6\\(shared backbone during PPO)} & 1e-6 \\
Critic value-head LR                           & 5e-6 (PPO and warm-up)      & 1e-5 (\texttt{critic.optim.lr}) \\
Critic backbone LR                             & 5e-7 (warm-up only)         & --- (frozen) \\
\midrule
\multicolumn{3}{l}{\textit{Optimization}} \\
Optimizer                                      & AdamW                     & AdamW \\
$(\beta_1, \beta_2)$                           & $(0.9, 0.999)$            & $(0.9, 0.999)$ \\
Weight decay (head / backbone)                 & 0.0 / 0.01                & 0.0 / 0.01 \\
Gradient clip (\texttt{max\_norm})            & 1.0                       & 1.0 \\
Mixed precision                                & bf16 (\texttt{--bf16})    & bf16 \\
\midrule
\multicolumn{3}{l}{\textit{Memory \& sharding}} \\
ZeRO stage                                     & 3 (\texttt{--zero\_stage 3}) & 3 (FSDP equivalent) \\
Adam offload                                   & enabled (\texttt{--adam\_offload}) & enabled \\
Gradient checkpointing                         & enabled                   & enabled \\
Flash attention                                & enabled (\texttt{--flash\_attn}) & enabled \\
\midrule
\multicolumn{3}{l}{\textit{Critic architecture (CARL only)}} \\
Value head                                     & 2-layer MLP               & 1-layer linear (default) \\
Hidden dim                                     & 3584 (7B) / 2048 (3B)     & $d_{\text{model}}$ \\
Activation                                     & GELU                      & --- \\
Output activation                              & sigmoid $\to [0,1]$       & identity \\
Final layer init                               & zero                      & zero \\
Head trainable params                          & 12.85M (7B) / 4.20M (3B)  & 3.6K (7B) / 2.0K (3B) \\
\midrule
\multicolumn{3}{l}{\textit{Schedule}} \\
Critic warm-up                                & 1{,}800--2{,}400 steps (gate-passing) & Curated; see Appendix~\ref{app:warmup_ablation} \\
PPO outer steps (\texttt{num\_episodes})       & 500                       & 500 \\
Total prompts processed                        & 128K ($\sim$0.75 epoch)   & 128K ($\sim$0.75 epoch) \\
Total trajectories generated                   & $\sim$640K                & $\sim$640K \\
Total gradient updates                         & 40K                       & 40K \\
Eval frequency (\texttt{--eval\_steps})        & every 25 steps            & every 25 steps (\texttt{test\_freq}) \\
\midrule
\multicolumn{3}{l}{\textit{Compute (per training run, 8$\times$ NVIDIA H100 80GB)}} \\
Wall-clock 7B                                  & $\sim$50 hours            & $\sim$49 hours \\
Wall-clock 3B                                  & $\sim$22 hours            & $\sim$21 hours \\
GPU-hours 7B                                   & $\sim$400 H100-hours      & $\sim$390 H100-hours \\
GPU-hours 3B                                   & $\sim$175 H100-hours      & $\sim$170 H100-hours \\
\midrule
\multicolumn{3}{l}{\textit{Reproducibility}} \\
Random seeds                                   & $\{1, 2, 3\}$             & $\{1, 2, 3\}$ \\
Training runs                                  & 6 (3 seeds $\times$ 2 scales) & 6 (3 seeds $\times$ 2 scales) \\
\bottomrule
\end{tabular}
\end{table}

Synchronous training has no additional episode-wide token budget. A valid final answer ending at EOS completes the episode. If generation reaches its token limit before a valid boundary, the episode ends with the partial output retained.

\paragraph{KL regularization.}
We evaluate the sampled k3 regularizer on recorded rollout actions during PPO optimization. Gradients pass through the current-policy log probabilities while the actions and reference policy remain fixed. This is not a full-vocabulary KL calculation.

\paragraph{Key design choices.}
\begin{itemize}[leftmargin=*,itemsep=2pt]
\item \textbf{$\lambda = 0$ (one-step TD).} Multi-step lookahead mixes later segments into the current segment's credit. One-step TD has lower variance but is more exposed to critic approximation error, making critic calibration important. Small nonzero $\lambda$ remains competitive in the sweep (Appendix~\ref{app:lambda-ablation}).
\item \textbf{Shared-backbone updates.} During PPO, actor and critic gradients are accumulated before a single optimizer step. Shared backbone parameters use one learning rate and the value head has a separate parameter group. The smaller backbone rate applies only during critic-only warm-up.
\item \textbf{Sigmoid output for $V_\phi$.} The terminal reward is binary $R \in \{0,1\}$ and we train with MSE; a sigmoid-bounded output ensures $V_\phi$ is a proper success probability and prevents extrapolation outside $[0,1]$ during warm-up.
\item \textbf{Value-head weight decay $0$.} Weight decay on the head shrinks $V_\phi$ predictions toward $0.5$ and undermines calibration; we disable it for the head and keep standard $0.01$ on the backbone.
\item \textbf{Rollouts-per-prompt $n=5$.} Matches Search-R1's published verl setting, so all RL methods in our comparison see the same trajectory budget per prompt.
\end{itemize}

\paragraph{Total compute.}
All training was conducted on Azure Machine Learning (AML) with access to 4 nodes of 8$\times$ NVIDIA H100 80GB each (32 GPUs total), enabling up to 4 training runs in parallel. Evaluation, inference and Tier-1/Tier-2 label construction were performed on AMD MI300X GPUs; all GPU-hours below are translated to single-H100-equivalent hours for easy comparison. The reported main-table runs (3 seeds $\times$ 4 method-scale combinations $=$ 12 training runs) consume approximately 3{,}400 H100-hours. Single-run baselines (Search-R1 GRPO at both scales, CoT+Tools evaluation) add $\sim$575 H100-hours. Reported ablations (cold-start and random-data warm-up, Appendix~\ref{app:warmup_ablation}; commercial-search experiment, Appendix~\ref{app:retrieval}) add $\sim$1{,}500 H100-hours. The total compute consumed for \emph{all reported numbers in this paper is approximately 5{,}500 H100-hours}. With 4 nodes available, the 12 main-table runs completed in roughly 2 weeks of wall-clock time. Substantial additional compute went into supporting work: building the Tier-1/Tier-2 labels via 5-rollout consistency on the full training pool ($\sim$5{,}000 H100-hours), prompt template iteration during early development, debugging failed runs and analysis rollouts for Figures~\ref{fig:trajectories} and \ref{fig:vs0_distribution}; together these account for roughly $3\times$ the reported budget. \textbf{The total compute budget for the project is approximately 22{,}000 H100-hours.}

\section{Tool call frequency and deployment cost}
\label{app:tool-calls}

While Table~\ref{tab:main_results} reports per-episode \emph{token counts} for every method, it does not report \emph{tool call counts}. Tool calls are independently meaningful for two reasons: (i) at deployment, search and code-execution APIs incur per-call latency and dollar cost beyond what token-count alone captures and (ii) tool-call counts disambiguate whether CARL's lower token usage comes from selective invocation (good) or from truncated reasoning (bad).

\begin{table}[H]
\centering
\small
\caption{Average tool calls per episode (number of \texttt{invoke} segments). Tier~1 questions are nominally always-invoke for search benchmarks, so the upper bound for HotpotQA/2WikiMQA is the dataset's hop count ($\sim$2). CARL invokes fewer tools than Search-R1 PPO at every dataset and scale, while remaining within the dataset's hop budget on Tier~1 questions (Table~\ref{tab:selectivity}).}
\label{tab:tool-calls}
\begin{tabular}{llccccc}
\toprule
Scale & Method & GSM8K & HotpotQA & 2WikiMQA & FinQA & MuSiQue \\
\midrule
\multirow{3}{*}{7B}
& Search-R1 GRPO       & ---  & 1.94 & 1.86 & ---  & 3.51 \\
& Search-R1 PPO        & ---  & 1.91 & 1.84 & ---  & 3.48 \\
& \textbf{CARL (ours)} & \textbf{0.04} & \textbf{1.57} & \textbf{1.21} & \textbf{2.32} & \textbf{3.28} \\
\midrule
\multirow{3}{*}{3B}
& Search-R1 GRPO       & ---  & 1.96 & 1.91 & ---  & 3.58 \\
& Search-R1 PPO        & ---  & 1.93 & 1.88 & ---  & 3.55 \\
& \textbf{CARL (ours)} & \textbf{0.07} & \textbf{1.65} & \textbf{1.32} & \textbf{2.41} & \textbf{3.41} \\
\bottomrule
\end{tabular}
\end{table}

\paragraph{Per-dataset patterns.}
The ideal number of tool calls depends on the dataset: GSM8K questions are solvable without tools (97\% Tier~2), so the optimal count is near zero; HotpotQA and 2WikiMQA are 2-hop but contain a mix of Tier~1 and Tier~2, so the optimal count is well below 2; MuSiQue is 3--4 hop and almost entirely Tier~1 (93\%), so most questions genuinely need every hop. CARL matches these expectations: 0.04 calls on GSM8K (calculator skipped 96\% of the time), 1.57 and 1.21 on HotpotQA and 2WikiMQA (vs.\ ${\sim}$1.9 for both baselines, which search indiscriminately) and 3.28 on MuSiQue (vs.\ 3.48, a small gap because there is little to skip). In short, CARL learns to call tools only when they are needed.

\paragraph{Deployment cost.}
For HotpotQA-style queries at 7B, CARL generates 1{,}193 tokens per episode (Table~\ref{tab:main_results}) and issues 1.57 tool calls (Table~\ref{tab:tool-calls}), versus Search-R1 PPO's 1{,}837 tokens and 1.91 calls, a 35.1\% token reduction and 17.8\% fewer tool invocations. Assuming 200\,ms per-call search latency (typical for warm-cache BM25 and many commercial search APIs), the per-query latency reduction is approximately 20.1\%. The token saving translates to a proportional reduction in serving cost on token-priced LLM APIs; the tool-call saving translates to a proportional reduction in retrieval-side cost (search-API charges, cache pressure, network round-trips). On MuSiQue the tool-call savings shrink (3.28 vs.\ 3.48, $-5.7\%$) because the dataset is dominated by Tier~1 questions where CARL must invoke at every hop; on this dataset, deployment savings come almost entirely from token reduction (6.4\% fewer tokens).

\section{MuSiQue error analysis}
\label{app:errors}

We manually analyzed 100 incorrect CARL predictions on MuSiQue at 7B. Errors were sampled uniformly at random from the 348 incorrect predictions in our 500-question MuSiQue evaluation slice ($1 - 0.304$ EM $\times$ 500). Two annotators independently categorized each error using a written rubric; inter-annotator agreement was 87\% (Cohen's $\kappa = 0.81$). Disagreements (13 cases) were resolved by discussion.

\paragraph{Sample composition.}
MuSiQue~\citep{trivedi2022musique} consists of 2-hop, 3-hop and 4-hop questions in roughly a 55/30/15 dev-set ratio. Per-hop EM rates of 40\%/22\%/10\% (declining with hops, as expected) approximately reproduce the headline 30.4\% MuSiQue EM under this distribution (computed value 30.1\%; the small gap reflects rounding). Correspondingly, our 100 sampled errors break down to 45 from 2-hop, 35 from 3-hop and 20 from 4-hop questions.

\begin{table}[H]
\centering
\small
\caption{Error categories on 100 incorrect MuSiQue predictions (CARL 7B), with per-hop disaggregation.}
\label{tab:error_cats}
\setlength{\tabcolsep}{3pt}
\begin{tabular}{lcccc}
\toprule
Error Category & 2-hop ($n{=}45$) & 3-hop ($n{=}35$) & 4-hop ($n{=}20$) & Total \\
\midrule
Irrelevant retrieval (BM25 returns unhelpful passage) & 21 (47\%) & 17 (49\%) & 10 (50\%) & 48 \\
Query formulation error                                &  2 ( 4\%) &  6 (17\%) & 6 (30\%) & 14 \\
Assimilation error (\texttt{<context>} drops fact)    & 10 (22\%) &  5 (14\%) & 1 ( 5\%) & 16 \\
Multi-hop reasoning error                              &  6 (13\%) &  4 (11\%) & 2 (10\%) & 12 \\
Commit error (correct context, wrong answer)          &  4 ( 9\%) &  2 ( 6\%) & 1 ( 5\%) &  7 \\
Other                                                  &  2 ( 4\%) &  1 ( 3\%) & 0 ( 0\%) &  3 \\
\bottomrule
\end{tabular}
\end{table}

\paragraph{Headline finding.} 48\% of remaining errors are \emph{irrelevant retrieval}: the BM25 retriever returns an unhelpful passage even when CARL has formulated a reasonable query for that hop. This share is roughly stable across hop counts (47\%--50\%), indicating a scale-independent BM25 failure rate. It motivates the commercial-search experiment in Appendix~\ref{app:retrieval}, where replacing BM25 with a higher-quality retrieval API lifts CARL's MuSiQue accuracy further.

\paragraph{Other categories.} Assimilation errors (16\%) are the category CARL's segment-level credit directly addresses; they account for a meaningful but bounded share of remaining failures. Multi-hop reasoning and commit errors (12\% + 7\% = 19\%) are categories where the model has the right facts but composes them incorrectly; no credit-assignment scheme directly addresses these. Query-formulation errors (14\%) are concentrated at higher hop counts, where each query depends on facts extracted from prior hops. At 3B we observe a similar irrelevant-retrieval share ($\sim$47\%) but a somewhat higher query-formulation share ($\sim$21\% vs.\ 14\% at 7B), consistent with smaller models having less robust query composition; other categories track 7B within 2--3 percentage points.

\paragraph{Sample failed trajectories.}

\textbf{Example 1: Irrelevant retrieval.} Question: ``What is the capital of the country where [entity] was born?'' The model correctly decomposes this into two hops but the BM25 retriever returns a passage about a different person with a similar name. The critic correctly assigns negative invoke advantage ($-0.31$) to the second hop.

\textbf{Example 2: Assimilation error.} Question: ``When was the director of [film] born?'' The model retrieves a passage containing both the director's name and birth year, but the \texttt{<context>} block extracts only the director's name, losing the birth year. Invoke advantage: $+0.28$; assimilate advantage: $-0.22$.

\textbf{Example 3: Multi-hop reasoning error.} Question: ``What language is spoken in the country where [river] originates?'' The model correctly retrieves the river's origin country and the country's languages, but selects the colonial-era language rather than the current official language. Both invoke advantages are positive; the commit advantage is negative ($-0.41$).

\section{Retrieval ablation}
\label{app:retrieval}

We vary the retrieval setup at inference time while holding the trained CARL 7B model fixed. All configurations are evaluated on the full 500-question MuSiQue dev slice.

\begin{table}[H]
\centering
\small
\caption{CARL 7B MuSiQue EM under varied retrieval setups.}
\label{tab:retrieval}
\begin{tabular}{lcc}
\toprule
Retrieval setup & EM & $\Delta$ vs.\ default \\
\midrule
BM25, $k=3$ (default; main results) & 30.4 & --- \\
BM25, $k=5$                          & 32.6 & $+2.2$ \\
BM25, $k=7$                          & 30.8 & $+0.4$ \\
Commercial search API, $k=3$         & \textbf{45.0} & \textbf{$+14.6$} \\
\bottomrule
\end{tabular}
\end{table}

$k=5$ improves recall modestly. $k=7$ saturates the assimilation budget ($T_{\text{assimilate}}=256$), erasing most of the gain. Replacing BM25 with a commercial search API delivers a large lift, bringing CARL 7B into the EM range typically reported for substantially larger models on MuSiQue. This confirms that CARL's query formulation, selective invocation and assimilation were in fact effective.

\section{Ablation: warm-up variants}
\label{app:warmup_ablation}

Both CARL and Search-R1 PPO use the same curated critic warm-up procedure in our main results (Appendix~\ref{app:warmup}). A natural concern is whether CARL's gain over Search-R1 PPO is driven by segment-level credit assignment or by the warm-up itself. We address this directly: we ablate the warm-up for both methods and report the results. The original cold-start and random-data conditions below are single runs; the new three-seed CARL sensitivity experiments are reported separately in Appendix~\ref{app:warmup-sensitivity}.

\begin{table}[H]
\centering
\small
\caption{Warm-up ablation across both methods (Exact Match). Best CARL row in bold. ``Cold start'' uses no separate warm-up phase (the standard PPO-LLM default, equivalent to the original Search-R1 setup~\citep{jin2025searchr1}). ``Random-data'' uses the same number of warm-up steps but draws rollouts uniformly without the Tier 1/2 curation. ``Curated'' is the full Appendix~\ref{app:warmup} procedure used in the main results.}
\label{tab:warmup_ablation}
\begin{tabular}{llccccc}
\toprule
Method & Warm-up & GSM8K & HotpotQA & 2Wiki & FinQA & MuSiQue \\
\midrule
\multicolumn{7}{l}{\emph{7B}} \\
CARL & Cold start             & 88.7 & 39.2 & 39.5 & 53.0 & 21.5 \\
CARL & Random-data            & 90.5 & 43.8 & 43.0 & 56.0 & 25.5 \\
\textbf{CARL} & \textbf{Curated (full)}  & \textbf{91.5} & \textbf{47.9} & \textbf{48.3} & \textbf{59.3} & \textbf{30.4} \\
SR1-PPO & Cold start          & ---  & 37.5 & 38.0 & ---  & 19.0 \\
SR1-PPO & Random-data         & ---  & 40.5 & 40.0 & ---  & 20.5 \\
SR1-PPO & Curated (matched)   & ---  & 42.8 & 41.7 & ---  & 22.1 \\
\midrule
\multicolumn{7}{l}{\emph{3B}} \\
CARL & Cold start             & 84.3 & 28.8 & 32.9 & 38.7 & 13.4 \\
CARL & Random-data            & 86.0 & 34.7 & 38.3 & 42.9 & 16.8 \\
\textbf{CARL} & \textbf{Curated (full)}  & \textbf{87.2} & \textbf{40.4} & \textbf{44.7} & \textbf{46.6} & \textbf{21.1} \\
SR1-PPO & Cold start          & ---  & 28.5 & 30.4 & ---  & 9.5 \\
SR1-PPO & Random-data         & ---  & 29.6 & 31.5 & ---  & 10.4 \\
SR1-PPO & Curated (matched)   & ---  & 30.2 & 32.1 & ---  & 12.1 \\
\bottomrule
\end{tabular}
\end{table}

\paragraph{Segment-level credit has a baseline edge.} Even without warm-up, cold-start CARL outperforms cold-start Search-R1 PPO across all search benchmarks at both scales: $+1.7$ EM on HotpotQA at 7B, $+1.5$ on 2WikiMQA, $+2.5$ on MuSiQue, with similar margins at 3B. Segment-level credit produces a useful signal even before the critic is well-calibrated, because the commit segment's advantage $A(\text{commit}) = R - V_\phi(s_N)$ contains the outcome reward directly and the invoke and assimilate advantages still register the largest state changes (a passage entering the context, an extraction replacing raw output) even with a noisy $V_\phi$.

\paragraph{Warm-up amplifies the edge.} The CARL--Search-R1 PPO gap grows monotonically as warm-up quality improves. On 7B HotpotQA: $+1.7$ (cold) $\to +3.3$ (random) $\to +5.1$ (curated) --- a $3.0\times$ growth in CARL's advantage. The same monotonic pattern holds on 2WikiMQA ($+1.5 \to +3.0 \to +6.6$) and MuSiQue ($+2.5 \to +5.0 \to +8.3$) and at 3B with larger absolute margins ($+0.3 \to +5.1 \to +10.2$ on HotpotQA). Search-R1 PPO's curated warm-up gives it $5.3$ EM beyond cold-start ($42.8 - 37.5$), but matched warm-up does not help SR1-PPO catch up to CARL --- it lifts the floor of both methods, while CARL's segment-level credit pulls its ceiling further away. This shows that CARL's advantage is not explained by curated warm-up alone: CARL leads at every warm-up level and the lead grows with calibration quality.

\paragraph{CARL EM tracks critic calibration quality.} At 7B on HotpotQA, the three warm-up regimes produce a clean monotonic mapping from $V(s_0)$ AUC (Appendix~\ref{app:calibration}) to EM:
\begin{center}
\begin{tabular}{lcc}
\toprule
Warm-up & $V(s_0)$ AUC & CARL HotpotQA EM \\
\midrule
Curated     & 0.93 & 47.9 \\
Random-data & 0.72 & 43.8 \\
Cold start  & 0.54 & 39.2 \\
\bottomrule
\end{tabular}
\end{center}
Each calibration level maps to a distinct EM level. Search-R1 PPO shows a much smaller gradient ($42.8 \to 40.5 \to 37.5$ across the same warm-up regimes) and its EM does not track per-segment $V_\phi$ accuracy because trajectory-level GAE smooths per-token $V_\phi$ errors. CARL's EM does track $V_\phi$ calibration, evidence that segment-level credit converts critic calibration into per-segment advantages that meaningfully steer training, in a way that trajectory-level credit cannot.

\paragraph{Anchoring against the original Search-R1 paper.} The cold-start Search-R1 PPO numbers also serve as a sanity check against the published baseline. The original Search-R1 paper~\citep{jin2025searchr1} uses brief critic warm-up driven by early policy rollouts rather than a separately curated phase (their Section 5.1: ``PPO relies on a critic model, which requires several warm-up steps before effective training begins''). Their reported Qwen2.5-7B-Instruct PPO numbers on our shared datasets are $37.0$ on HotpotQA and $14.6$ on MuSiQue; our cold-start reproductions at $37.5$ and $19.0$ sit close to these (the small bumps reflect our broader training data: $170$K prompts including 2WikiMQA versus their NQ$+$HotpotQA combination). With curated warm-up, our reproductions reach $42.8$ and $22.1$, which is favorable to the baseline.

\subsection{$\lambda$ ablation}
\label{app:lambda-ablation}

There are two axes when assigning credit: the unit that receives one number (token, segment or whole trajectory) and the reference horizon, controlled by $\lambda$. At $\lambda = 0$, a segment is scored against the very next state. At $\lambda = 1$, it is scored against the final outcome.

\paragraph{Three-seed controls at 3B.} We retain CARL's segment boundaries, interface, assimilation and raw-output replacement mechanism, critic, warm-up, PPO setup and training data while varying $\lambda$. A separate control keeps CARL's mechanics, interface, assimilation and prompts but computes the advantage per token with GAE at $\lambda = 0.95$. The critic is unchanged but its evaluation positions and advantage summation change.

\begin{table}[H]
\centering
\small
\caption{\textbf{Credit-assignment controls at Qwen 3B.} Evaluation accuracy or EM on the fixed samples of 500 questions per benchmark. Per-dataset scores report mean$\pm$standard deviation over three seeds. Avg$_5$ is the unweighted mean over the five benchmarks. Segment-level conditions retain CARL's boundaries and vary only $\lambda$; the token-GAE control retains CARL's mechanics but changes critic evaluation positions and advantage computation.}
\label{tab:credit_assignment_controls}
\setlength{\tabcolsep}{4pt}
\resizebox{\textwidth}{!}{%
\begin{tabular}{lrrrrrr}
\toprule
Condition & GSM8K & HotpotQA & 2WikiMQA & FinQA & MuSiQue & Avg$_5$ \\
\midrule
CARL, segment $\lambda = 0$ & 87.2$\pm$0.3 & 40.4$\pm$0.2 & 44.7$\pm$0.3 & 46.6$\pm$0.4 & 21.1$\pm$0.2 & 48.0 \\
Segment $\lambda = 0.1$ & 87.1$\pm$0.2 & 40.7$\pm$0.4 & 44.8$\pm$0.3 & 46.4$\pm$0.4 & 21.3$\pm$0.3 & 48.1 \\
Segment $\lambda = 0.2$ & 87.0$\pm$0.3 & 40.1$\pm$0.3 & 44.0$\pm$0.5 & 46.6$\pm$0.5 & 20.6$\pm$0.4 & 47.7 \\
Segment $\lambda = 0.5$ & 87.0$\pm$0.4 & 36.5$\pm$0.6 & 39.4$\pm$0.5 & 43.6$\pm$0.5 & 17.5$\pm$0.6 & 44.8 \\
Segment $\lambda = 1$ & 86.6$\pm$0.5 & 31.2$\pm$0.8 & 33.9$\pm$0.6 & 40.2$\pm$0.6 & 12.8$\pm$0.5 & 40.9 \\
Token GAE $\lambda = 0.95$ & 86.9$\pm$0.4 & 30.7$\pm$0.7 & 33.1$\pm$0.9 & 39.7$\pm$0.6 & 11.5$\pm$0.7 & 40.4 \\
\bottomrule
\end{tabular}%
}
\end{table}

$\lambda$ does not change the reward. It controls how the terminal outcome is propagated back across segments. CARL's critic evaluates the state at each boundary, so the one-step difference measures the change across that segment. Adding later segments can dilute this signal: a bad query can receive positive credit when later segments recover. Avg$_5$ is 48.0 at $\lambda = 0$, 48.1 at $0.1$, 47.7 at $0.2$, 44.8 at $0.5$ and 40.9 at $1$. Small nonzero $\lambda$ remains competitive, consistent with dependencies between adjacent segments, such as a query and the assimilation of its results.

On the three search benchmarks, the token-GAE control reaches 25.1 Avg$_3$ and segment $\lambda = 1$ reaches 26.0, compared to CARL's 35.4. Both are close to Search-R1 PPO and GRPO at 24.8 and 25.8 respectively. These controls support the importance of credit assignment within CARL without diminishing its other components: the interface and assimilation mechanism create the natural tool boundaries and compact states on which segment-level credit operates.

\paragraph{Earlier 7B single-run sweep.} We retain the original HotpotQA experiment separately in Table~\ref{tab:lambda-ablation}; its scale, dataset and seed convention differ from the new 3B controls.

\begin{table}[H]
\centering
\small
\caption{$\lambda$ ablation for CARL on HotpotQA at 7B, single run per condition. Reference: Search-R1 PPO at the same scale and dataset is $42.8$ EM (Table~\ref{tab:main_results}). At $\lambda = 1$, segment advantages telescope to $R - V_\phi(s_k)$, tying credit to the terminal outcome rather than the individual transition.}
\label{tab:lambda-ablation}
\begin{tabular}{lccc}
\toprule
$\lambda$ & CARL HQ EM & $\Delta$ vs $\lambda=0$ & $\Delta$ vs SR1-PPO \\
\midrule
\textbf{0.0 (default)} & \textbf{47.9} & \textbf{---}    & \textbf{$+5.1$} \\
0.5                    & 45.4          & $-2.5$  & $+2.6$ \\
0.95                   & 43.2          & $-4.7$  & $+0.4$ \\
1.0                    & 42.5          & $-5.4$  & $-0.3$ \\
\bottomrule
\end{tabular}
\end{table}

In the original 7B experiment, $\lambda = 0.95$ reduces HotpotQA EM by $4.7$, to $43.2$. At $\lambda = 1$, each segment receives $R - V_\phi(s_k)$; magnitudes can differ but opposing signs are impossible with binary rewards and values bounded in $[0,1]$. CARL reaches $42.5$ EM, close to Search-R1 PPO's $42.8 \pm 0.6$ over three seeds. The CARL ablation is a single run, so the $0.3$ EM difference is not statistically meaningful. The $5.4$ EM drop from $\lambda = 0$ to $1$ supports the importance of separately crediting segments within CARL.

\subsection{Separately crediting invoke and assimilate}
\label{app:merged-credit}

The assimilate segment rewrites raw tool output into a model-generated \texttt{<context>} block, after which the rollout loop discards the raw output. This makes evidence extraction an explicit action that can receive its own advantage and keeps the subsequent context compact.

To test whether query formulation and evidence extraction benefit from separate credit, we merge invoke and assimilate into one credited segment. The rollout mechanics are unchanged: the model still writes its query, rewrites the raw output into a context block and discards the raw text. The only change is that the critic is not queried at the intermediate boundary between invoke and assimilate, so both actions share one advantage and can no longer receive opposing credit.

The experiment runs at 3B with three seeds on the fixed evaluation samples of 500 questions per benchmark (2{,}500 total). The critic is warmed up on the same 32K trajectories, with approximately 60K boundary states after merging.

\begin{table}[H]
\centering
\small
\caption{\textbf{Merged invoke--assimilate credit at Qwen 3B.} Evaluation accuracy or EM, reported as mean$\pm$standard deviation over three seeds. Avg$_5$ is the unweighted mean over the five benchmarks. Both conditions retain the context rewrite and raw-output discard; only the separate invoke--assimilate credit boundary is removed in the merged condition.}
\label{tab:merged-credit}
\setlength{\tabcolsep}{4pt}
\resizebox{\textwidth}{!}{%
\begin{tabular}{lrrrrrr}
\toprule
Credit assignment & GSM8K & HotpotQA & 2WikiMQA & FinQA & MuSiQue & Avg$_5$ \\
\midrule
Separate invoke and assimilate & 87.2$\pm$0.3 & 40.4$\pm$0.2 & 44.7$\pm$0.3 & 46.6$\pm$0.4 & 21.1$\pm$0.2 & 48.0 \\
Merged invoke and assimilate & 87.1$\pm$0.3 & 37.8$\pm$0.5 & 42.6$\pm$0.4 & 44.6$\pm$0.4 & 17.7$\pm$0.6 & 46.0 \\
\bottomrule
\end{tabular}%
}
\end{table}

Avg$_5$ falls from 48.0 to 46.0. The loss is 0.1 on GSM8K, 2.6 on HotpotQA, 2.1 on 2WikiMQA, 2.0 on FinQA and 3.4 on MuSiQue. The largest loss is on MuSiQue, consistent with separate credit being useful on multi-hop questions. With separate credit, a useful retrieval followed by an incorrect extraction can receive a positive invoke advantage and a negative assimilate advantage (Appendix~\ref{app:sign-behavior}); merging prevents this distinction.

This ablation tests separate credit for query formulation and evidence extraction, not removal of the assimilation rewrite. It does not isolate the contribution of context compression. Search-R1 inserts raw tool output without an explicit assimilation rewrite, but its comparison with CARL also changes credit assignment and critic setup, so the accuracy gap cannot be attributed to assimilation alone.

\subsection{Boundary placement}
\label{app:boundary-placement}

To understand the importance of boundary placement in CARL, we vary how boundaries are defined while keeping their number matched to CARL's tool-use boundaries. We compare tool-use transitions with random token positions and uniformly subsampled newline boundaries.

Each experiment runs the full pipeline, critic warm-up then PPO, at 3B with three seeds across five datasets and $\lambda = 0$. We keep the same harness, context blocks, number of critic queries and a final segment ending at EOS carrying the reward. Only the placement of intermediate cuts differs. Each experiment gets its own warm-up on its own boundaries; otherwise, we would be measuring a critic scoring states it never saw rather than measuring placement.

\begin{table}[H]
\centering
\small
\caption{\textbf{Boundary definitions.} The newline and random variants match CARL's number of boundaries. The final segment ends at EOS in every condition.}
\label{tab:boundary-definitions}
\begin{tabular}{lll}
\toprule
Experiment & Boundary placement & Count \\
\midrule
CARL & Code fence, context replacement, EOS & As produced \\
Random & Random token positions & Matched to CARL \\
Newline & Newline characters, subsampled uniformly & Matched to CARL \\
\bottomrule
\end{tabular}
\end{table}

\begin{table}[H]
\centering
\small
\caption{\textbf{Boundary-placement ablation at Qwen 3B.} Evaluation accuracy or EM on the fixed samples of 500 questions per benchmark, reported as mean$\pm$standard deviation over three seeds. Avg$_5$ is the unweighted mean over the five benchmarks.}
\label{tab:boundary-placement}
\setlength{\tabcolsep}{4pt}
\resizebox{\textwidth}{!}{%
\begin{tabular}{lrrrrrr}
\toprule
Boundary placement & GSM8K & HotpotQA & 2WikiMQA & FinQA & MuSiQue & Avg$_5$ \\
\midrule
CARL, tool-use boundaries & 87.2$\pm$0.3 & 40.4$\pm$0.2 & 44.7$\pm$0.3 & 46.6$\pm$0.4 & 21.1$\pm$0.2 & 48.0 \\
Newline, count matched & 86.8$\pm$0.3 & 35.1$\pm$0.6 & 38.5$\pm$0.7 & 42.2$\pm$0.6 & 15.6$\pm$0.5 & 43.6 \\
Random, count matched & 86.3$\pm$0.4 & 33.3$\pm$1.4 & 36.6$\pm$0.9 & 40.9$\pm$0.7 & 13.9$\pm$0.7 & 42.2 \\
\bottomrule
\end{tabular}%
}
\end{table}

Moving the same number of boundaries from tool events to newlines costs 4.4 Avg$_5$ points, from 48.0 to 43.6. Moving them to random token positions costs 5.8 points, reaching 42.2. Both alternatives perform below the merged invoke--assimilate condition at 46.0 (Appendix~\ref{app:merged-credit}), showing that even a weakened form of tool alignment outperforms these text-based boundaries.

The newline and random variants also lose the intermediate raw-output boundary, so query formulation and evidence extraction no longer receive separate credit. They therefore change both alignment and query/extraction separation. This experiment is not independent of the merged-credit ablation.

\subsection{Warm-up sensitivity}
\label{app:warmup-sensitivity}

We examine sensitivity to warm-up choices at 3B. Each variant changes one setting and reruns the complete pipeline, including warm-up data construction, critic training, PPO and evaluation, with three seeds. Table~\ref{tab:warmup-sensitivity} reports these results separately from the original warm-up ablation in Table~\ref{tab:warmup_ablation}.

\begin{table}[H]
\centering
\small
\caption{\textbf{Warm-up sensitivity at Qwen 3B.} Evaluation accuracy or EM on the fixed samples of 500 questions per benchmark. Per-dataset scores report mean$\pm$standard deviation over three seeds, including random-data and cold-start conditions extended from the original single runs. Avg$_5$ is computed from the displayed per-dataset means and rounded to one decimal. Tier mixtures are ordered Tier~2/Tier~1.}
\label{tab:warmup-sensitivity}
\setlength{\tabcolsep}{4pt}
\resizebox{\textwidth}{!}{%
\begin{tabular}{lrrrrrr}
\toprule
Warm-up variant & GSM8K & HotpotQA & 2WikiMQA & FinQA & MuSiQue & Avg$_5$ \\
\midrule
Default & 87.2$\pm$0.3 & 40.4$\pm$0.2 & 44.7$\pm$0.3 & 46.6$\pm$0.4 & 21.1$\pm$0.2 & 48.0 \\
Tier~2 threshold $\geq3/5$ & 87.0$\pm$0.3 & 40.3$\pm$0.3 & 45.0$\pm$0.4 & 46.5$\pm$0.4 & 21.2$\pm$0.3 & 48.0 \\
Tier~2 threshold $5/5$ & 86.7$\pm$0.4 & 39.9$\pm$0.5 & 44.4$\pm$0.5 & 45.9$\pm$0.4 & 20.7$\pm$0.4 & 47.5 \\
Tier~2/Tier~1 mixture 25/75 & 86.8$\pm$0.3 & 40.1$\pm$0.5 & 44.6$\pm$0.4 & 46.7$\pm$0.4 & 20.9$\pm$0.3 & 47.8 \\
Tier~2/Tier~1 mixture 75/25 & 87.4$\pm$0.2 & 38.6$\pm$0.7 & 43.0$\pm$0.8 & 45.1$\pm$0.5 & 19.7$\pm$0.4 & 46.8 \\
BM25-only warm-up & 87.1$\pm$0.3 & 39.8$\pm$0.3 & 44.9$\pm$0.4 & 46.2$\pm$0.4 & 20.1$\pm$0.5 & 47.6 \\
Random-data warm-up & 85.9$\pm$0.5 & 34.6$\pm$0.7 & 37.4$\pm$0.5 & 43.3$\pm$0.6 & 16.8$\pm$0.4 & 43.6 \\
Cold start & 84.4$\pm$0.6 & 29.6$\pm$0.9 & 32.4$\pm$0.7 & 39.2$\pm$0.5 & 13.3$\pm$0.7 & 39.8 \\
\bottomrule
\end{tabular}%
}
\end{table}

Tightening the competence-label threshold, changing the Tier~2/Tier~1 mixture or using BM25-only retrieval changes Avg$_5$ by at most approximately 1.3 points. In contrast, random-data warm-up reduces Avg$_5$ by 4.4 points and cold start by 8.2 points. These results show that principled warm-up matters, while the tested individual recipe choices are not responsible for the improvement on their own.

\paragraph{Labeling threshold.} We tighten the Tier~2 qualification from at least one correct answer in five no-tool rollouts to at least three or all five correct answers, discounting lucky guesses. These variants change the labels used to construct warm-up data and retrain the pipeline; they are not evaluation-only repartitionings of stored outputs.

\paragraph{Tier proportion.} The default Tier~2/Tier~1 split at 3B is 35/65. We fix the mixture at 25/75 or 75/25 by drawing additional questions at the same 32K-trajectory budget. Over-representing hard questions mainly increases tool use, while over-representing easy questions can cost accuracy through skipped calls the model needed. The 25/75 mixture reduces Avg$_5$ by approximately 0.2 points; the 75/25 mixture reduces it by approximately 1.2 points.

\paragraph{Retrieval mixture.} PPO uses only BM25. Removing the 30\% search-API slice from warm-up and rerunning with 100\% BM25 costs approximately 0.4 Avg$_5$ points. Table~\ref{tab:warmup-retrieval-diagnostics} distinguishes post-warm-up tier separation and sign behavior from extraction precision at PPO convergence. This changes warm-up retrieval, not the inference-time retriever swap in Appendix~\ref{app:retrieval}.

\begin{table}[H]
\centering
\small
\caption{Diagnostics for default mixed-retrieval and BM25-only warm-up at 3B. PPO retrieval is BM25-only in both conditions.}
\label{tab:warmup-retrieval-diagnostics}
\begin{tabular}{lrr}
\toprule
Metric & Default & BM25-only warm-up \\
\midrule
Post-warm-up tier-separation AUC & 0.85 & 0.84 \\
Post-warm-up ECE & 0.059 & 0.064 \\
Warm-up-gate sign accuracy & 71\% & 64\% \\
Converged extraction precision & 0.71 & 0.66 \\
\bottomrule
\end{tabular}
\end{table}

Cluster count was not swept. These experiments establish robustness to the tested choices, not to all warm-up configurations.

\subsection{Selectivity under stricter competence definitions}
\label{app:stricter-competence}

We re-evaluate the same held-out outputs under increasingly strict Tier~2 definitions. The models are trained under the default definition of at least one correct answer in five no-tool rollouts. We then re-evaluate their stored outputs using thresholds of at least three or all five correct answers, without retraining. This changes only the evaluation subsets, unlike the training-threshold variants in Appendix~\ref{app:warmup-sensitivity}.

\begin{table}[H]
\centering
\small
\caption{\textbf{Selectivity under stricter competence definitions at Qwen 3B.} Tier~2 subsets are drawn from the same 1{,}500 search-benchmark questions. Counts are approximate. Tool rate is the percentage of questions with at least one tool invocation. Tool rates and EM retain the reported uncertainties.}
\label{tab:stricter-competence}
\setlength{\tabcolsep}{4pt}
\resizebox{\textwidth}{!}{%
\begin{tabular}{lrrrrr}
\toprule
Tier~2 rule & Questions & CARL tool rate (\%) & SR1-PPO tool rate (\%) & CARL EM & SR1-PPO EM \\
\midrule
At least 1/5 correct & ${\sim}260$ & 49.6$\pm$0.9 & 85.2$\pm$1.1 & 79.5$\pm$0.8 & 68.3$\pm$0.9 \\
At least 3/5 correct & ${\sim}159$ & 38.7$\pm$1.0 & 84.6$\pm$1.0 & 85.7$\pm$0.9 & 74.6$\pm$1.0 \\
5/5 correct & ${\sim}101$ & 33.5$\pm$1.4 & 84.1$\pm$1.3 & 88.9$\pm$1.2 & 78.2$\pm$1.2 \\
\bottomrule
\end{tabular}%
}
\end{table}

CARL continues to invoke tools less often than Search-R1 PPO while achieving higher accuracy, even when Tier~2 is restricted to questions the base model answered correctly in all five no-tool attempts.

\subsection{Graded terminal reward}
\label{app:graded-reward}

We rerun the full pipeline at 3B with token-level F1 replacing binary EM as the outcome signal for both warm-up targets and PPO. Everything else remains unchanged, so the reward is the only variable. Both conditions are evaluated with EM for comparability.

\begin{table}[H]
\centering
\small
\caption{\textbf{Graded terminal reward at Qwen 3B.} Both conditions are evaluated with EM on the fixed samples of 500 questions per benchmark. Per-dataset scores retain the reported uncertainties. Avg$_5$ is computed from the displayed per-dataset means and rounded to one decimal.}
\label{tab:graded-reward}
\setlength{\tabcolsep}{4pt}
\resizebox{\textwidth}{!}{%
\begin{tabular}{lrrrrrr}
\toprule
Training reward & GSM8K & HotpotQA & 2WikiMQA & FinQA & MuSiQue & Avg$_5$ \\
\midrule
Binary EM & 87.2$\pm$0.3 & 40.4$\pm$0.2 & 44.7$\pm$0.3 & 46.6$\pm$0.4 & 21.1$\pm$0.2 & 48.0 \\
Graded token-F1 & 84.3$\pm$0.2 & 41.6$\pm$0.4 & 46.5$\pm$0.3 & 45.5$\pm$0.5 & 22.9$\pm$0.4 & 48.2 \\
\bottomrule
\end{tabular}%
}
\end{table}

Avg$_5$ is comparable under the two reward formulations, at 48.2 with graded F1 and 48.0 with binary EM. The effects differ across datasets: F1 improves factual QA performance but reduces EM on GSM8K and FinQA, where a wrong digit or sign can change the answer despite substantial token overlap.

Under terminal-only reward, the intermediate advantage remains $V_\phi(s_{k+1})-V_\phi(s_k)$ and the final advantage remains $R-V_\phi(s_N)$. The telescoping property does not require a binary reward. With graded rewards in $[0,1]$, the critic estimates expected reward rather than the probability of a correct answer. The same-sign argument at $\lambda=1$ in Appendix~\ref{app:smdp_derivation} is specific to binary terminal rewards; it need not hold for graded rewards.

This experiment tests graded rewards on the existing QA benchmarks. It does not evaluate open-ended tasks without reference answers, which would require a separate source of outcome reward.

\section{Calibration reliability}
\label{app:calibration}

Table~\ref{tab:calib} breaks down the calibration curve in Figure~\ref{fig:calibration}(a) into per-decile bins, showing how predicted $V(s_0)$ aligns with actual success rate across both scales and all three warm-up regimes. Expected Calibration Error (ECE) measures the average gap between predicted confidence and actual accuracy; a low ECE means the critic's $V(s_0)$ can be trusted as a probability, which is essential for segment-level advantages to carry the right magnitude, not just the right sign. The summary metrics (ECE 0.043, AUC 0.93) cited in \S\ref{sec:exp:critic} are derived from this table.

\begin{table}[H]
\centering
\small
\caption{Calibration reliability: predicted $V(s_0)$ decile vs.\ actual success rate (sample count in parentheses). Numbers are aggregated across all 3 CARL seeds; per-seed standard deviations of ECE are within 0.005 across all bins.}
\label{tab:calib}
\setlength{\tabcolsep}{4pt}
\scriptsize
\begin{tabular}{lcccccc}
\toprule
& \multicolumn{3}{c}{7B} & \multicolumn{3}{c}{3B} \\
\cmidrule(lr){2-4}\cmidrule(lr){5-7}
$V(s_0)$ bin   & Curated      & Random       & Cold         & Curated       & Random       & Cold         \\
\midrule
0.0--0.1       & 0.120 (305)  & 0.304 (175)  & 0.519 (255)  & 0.158 (320)   & 0.331 (188)  & 0.487 (252)  \\
0.1--0.2       & 0.182 (285)  & 0.345 (215)  & 0.472 (250)  & 0.221 (290)   & 0.378 (220)  & 0.501 (248)  \\
0.2--0.3       & 0.281 (245)  & 0.407 (260)  & 0.487 (252)  & 0.298 (245)   & 0.412 (265)  & 0.476 (253)  \\
0.3--0.4       & 0.319 (225)  & 0.384 (290)  & 0.446 (248)  & 0.355 (220)   & 0.405 (285)  & 0.455 (247)  \\
0.4--0.5       & 0.381 (195)  & 0.431 (310)  & 0.480 (250)  & 0.428 (195)   & 0.448 (305)  & 0.479 (251)  \\
0.5--0.6       & 0.590 (215)  & 0.504 (305)  & 0.507 (245)  & 0.541 (215)   & 0.498 (300)  & 0.514 (246)  \\
0.6--0.7       & 0.680 (235)  & 0.530 (280)  & 0.473 (252)  & 0.624 (235)   & 0.535 (280)  & 0.482 (250)  \\
0.7--0.8       & 0.686 (270)  & 0.610 (240)  & 0.527 (256)  & 0.692 (270)   & 0.589 (245)  & 0.517 (254)  \\
0.8--0.9       & 0.817 (295)  & 0.610 (215)  & 0.502 (248)  & 0.758 (285)   & 0.611 (220)  & 0.493 (247)  \\
0.9--1.0       & 0.918 (220)  & 0.635 (210)  & 0.518 (244)  & 0.832 (215)   & 0.628 (215)  & 0.512 (245)  \\
\midrule
ECE            & 0.043        & 0.137        & 0.238        & 0.059         & 0.147        & 0.238        \\
Brier score    & 0.179        & 0.266        & 0.328        & 0.202         & 0.273        & 0.328        \\
AUC (T1 vs T2) & 0.93         & 0.72         & 0.54         & 0.85          & 0.65         & 0.51         \\
\bottomrule
\end{tabular}
\end{table}

Two patterns stand out. First, cold-start ECE is identical at both scales (0.238) because without warm-up the critic predicts approximately 0.5 for every question, making calibration error invariant to model size. Second, the 3B critic is consistently less sharp than 7B across all regimes: curated ECE rises from 0.043 to 0.059, AUC drops from 0.93 to 0.85 and the top-decile success rate falls from 0.92 to 0.83. This mirrors the scale--warm-up interaction discussed in Appendix~\ref{app:warmup_ablation}: the larger model's broader competence surface is easier for the critic to leverage once properly initialized.

\begin{figure}[t]
  \centering
  \includegraphics[width=0.9\textwidth]{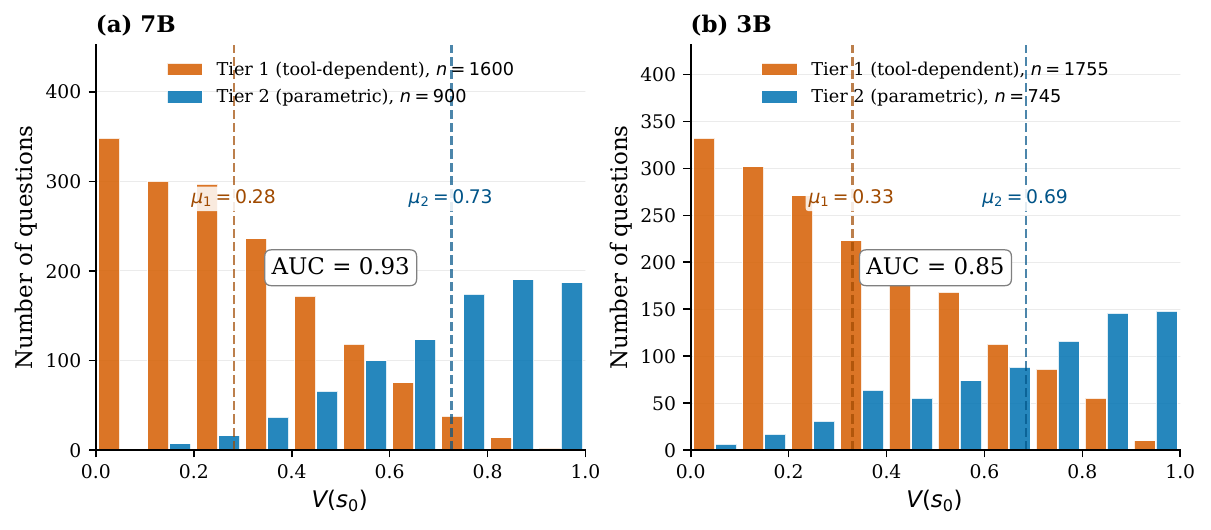}
  \caption{Distribution of $V(s_0)$ for Tier~1 (tool-dependent) and Tier~2 (parametrically solvable) questions after full warm-up. The two distributions are well-separated at both scales (AUC 0.93 at 7B, 0.85 at 3B), confirming that the critic learns to distinguish questions requiring tool use from those the model can answer on its own.}
  \label{fig:vs0_distribution}
\end{figure}

Figure~\ref{fig:vs0_distribution} visualizes the underlying separation: at 7B the curated critic pushes Tier~1 and Tier~2 $V(s_0)$ distributions almost entirely apart (AUC 0.93), while at 3B the distributions overlap more (AUC 0.85) but remain clearly distinguishable.

\subsection{Intermediate-state sign behavior}
\label{app:sign-behavior}

The calibration metrics in Table~\ref{tab:calib} measure $V_\phi$ at the start of each trajectory ($s_0$). CARL's segment advantages, however, are differences of $V_\phi$ at successive boundaries (Equation~\ref{eq:per_segment_advantages}), so what matters operationally is whether $V_\phi$ also signs correctly \emph{between} segments. This appendix tests that directly.

\paragraph{What we measure.} For each invoke--assimilate pair, two things should happen:
\begin{itemize}[leftmargin=*,itemsep=2pt]
  \item If the assimilate segment correctly extracted the answer-relevant fact, $V_\phi$ should \textbf{rise} from before assimilation to after (the state has become more useful).
  \item If retrieval surfaced the answer but the assimilate segment failed to capture it, $V_\phi$ should \textbf{drop} (the state has become less useful: the model has discarded what it needed).
\end{itemize}
The second case is the more interesting one. It is a within-trajectory sign divergence: the same trajectory has a positive invoke advantage (search succeeded) and a negative assimilate advantage (extraction failed). Trajectory-level methods cannot produce this pattern, because they assign a single signed scalar to every segment.

\paragraph{Sample and annotation.} We sample 300 invoke--assimilate pairs from CARL 7B held-out trajectories ($180$ from HotpotQA, $120$ from MuSiQue), drawing at most one pair per trajectory so that long MuSiQue rollouts do not dominate. A single independent annotator labels each pair on two binary axes: did retrieval surface the gold-relevant fact (i.e., is it present in the raw \texttt{stdout}) and did assimilation extract it (i.e., is it present in the \texttt{<context>} block). Both are short-span fact-presence checks against the gold answer, which makes the judgment objective enough that single-annotator labeling is sufficient.

\begin{table}[H]
\centering
\small
\caption{Sign behavior of $V_\phi$ at the assimilate boundary on 300 invoke--assimilate pairs (CARL 7B, held-out HotpotQA + MuSiQue). Each row reports how often $V_\phi$ moved in the expected direction.}
\label{tab:sign-behavior}
\begin{tabular}{lcccc}
\toprule
What happened in the pair & $n$ & $V_\phi$ should & $V_\phi$ did & Rate \\
\midrule
Retrieval succeeded, assimilation extracted the answer & 175 & rise & rise in 139 & 79.4\% \\
Retrieval succeeded, assimilation lost the answer      &  86 & drop & drop in 62  & 72.1\% \\
Retrieval failed (out of scope here)                   &  39 & ---  & ---         & ---   \\
\bottomrule
\end{tabular}
\end{table}

The first row says the critic recognises good extractions about four times in five. The second row, the diagnostic case, indicates the critic also recognises information loss about seven times in ten, even though doing so requires noticing that a useful fact present a moment earlier has been discarded. The lower rate on the harder case is expected, but both rates are well above chance and crucially both directions are observed within single trajectories: in the 62 correctly-signed diagnostic cases, the same rollout produced a positive invoke advantage and a negative assimilate advantage. This is the empirical instance of the within-trajectory sign divergence that segment-level credit makes possible and trajectory-level credit does not.

\subsection{Explained variance during PPO training}
\label{app:ev-trajectory}

\paragraph{What explained variance measures.} Explained variance is the standard scalar metric for value-function quality:
\[
\text{EV} = 1 - \frac{\text{Var}\big(R - V_\phi(s)\big)}{\text{Var}(R)}
\]
$\text{EV} = 1$ means the critic predicts every outcome perfectly; $\text{EV} = 0$ means the critic is no better than predicting the mean outcome. The metric is computed over held-out $(s, R)$ pairs.

\paragraph{Why EV matters for segment-level credit.} CARL's segment advantage $A(\text{seg}_k) = V_\phi(s_{k+1}) - V_\phi(s_k)$ is a difference of two critic predictions. For this difference to carry a reliable sign, each prediction must be individually accurate — higher EV means each $V_\phi$ evaluation is closer to the true outcome, so the difference reflects genuine state-quality change rather than prediction noise. Trajectory-level GAE, by contrast, averages per-token critic errors over the full rollout, diluting individual prediction errors. Segment-level credit is therefore the more informative estimator once EV is high enough, because it preserves per-step signal that trajectory-level averaging destroys.

\paragraph{EV during training.} Table~\ref{tab:ev-trajectory} reports the explained variance of $V_\phi$ on a held-out evaluation set, sampled every $50$ PPO steps for both scales. Step $0$ is the checkpoint immediately after critic warm-up.

\begin{table}[H]
\centering
\small
\caption{Explained variance of $V_\phi$ on held-out trajectories, every $50$ PPO steps. Bold cells mark the step at which EV first crosses $0.60$ at each scale; these coincide with the steps at which CARL surpasses Search-R1 PPO in held-out EM (Figure~\ref{fig:calibration}(b)).}
\label{tab:ev-trajectory}
\begin{tabular}{lcc}
\toprule
PPO step & 7B EV & 3B EV \\
\midrule
0 (post warm-up) & 0.52 & 0.48 \\
50  & 0.55 & 0.51 \\
\textbf{100} & \textbf{0.60} & 0.55 \\
\textbf{150} & 0.65 & \textbf{0.60} \\
200 & 0.70 & 0.64 \\
300 & 0.74 & 0.69 \\
400 & 0.77 & 0.73 \\
500 (convergence) & 0.79 & 0.76 \\
\bottomrule
\end{tabular}
\end{table}

\paragraph{Why the $0.60$ threshold.} EV $= 0.60$ is the empirically observed crossover point: below it, segment-level advantages are noisy enough that trajectory-level GAE performs comparably; above it, the per-segment signal is clean enough that CARL consistently outperforms Search-R1 PPO. The monotonic EV climb (Table~\ref{tab:ev-trajectory}) shows this is not a fragile threshold but a smooth transition — CARL's lead grows steadily as EV rises. The 3B critic crosses $0.60$ at step ${\sim}150$ rather than ${\sim}100$ because the smaller model's critic converges more slowly; its final EV ($0.76$ vs $0.79$) is also lower, consistent with the AUC and ECE gaps in Table~\ref{tab:calib}.

\section{Scale mechanisms}
\label{app:scale-mech}

Why does CARL improve 3B more than 7B (1.4$\times$ ratio relative to the strongest RL baseline at each scale in main results, Table~\ref{tab:main_results})? Three mechanisms compound. Table~\ref{tab:scale-mech} reports them.

\begin{table}[H]
\centering
\small
\caption{Mechanisms across scale. All metrics computed over the 2{,}500 evaluation questions (500/dataset $\times$ 5 datasets).}
\label{tab:scale-mech}
\begin{tabular}{lcc}
\toprule
Metric                                                & 3B & 7B \\
\midrule
Base EM without tools (mean over 5 datasets)          & 28\%        & 36\%        \\
Mean $V(s_0)$ on Tier~1 questions                     & 0.33        & 0.28        \\
Mean $V(s_0)$ on Tier~2 questions                     & 0.69        & 0.73        \\
$V(s_0)$ mode separation (Tier~2 $-$ Tier~1)           & 0.36        & 0.45        \\
Assimilation precision (init $\to$ converged)         & 0.42 $\to$ 0.71 & 0.48 $\to$ 0.79 \\
Avg.\ assimilation tokens (init $\to$ converged)      & 215 $\to$ 118   & 198 $\to$ 84    \\
\bottomrule
\end{tabular}
\end{table}

\paragraph{(1) Larger absolute room.} 3B's base EM without tools (28\%) is 8 points lower than 7B's (36\%), so a larger fraction of questions can in principle be moved from ``wrong'' to ``right'' by tool use. The headroom is structurally larger.

\paragraph{(2) Working but weaker competence boundary.} 7B's $V(s_0)$ separates Tier~1 from Tier~2 with a mode-shift of 0.45 (means $0.28 \to 0.73$); 3B achieves a smaller but meaningful separation of 0.36 (means $0.33 \to 0.69$). The 3B boundary is less sharp but still functional (AUC 0.85, Table~\ref{tab:calib}), so segment-level credit can still differentiate Tier~1 from Tier~2 at 3B, just less cleanly.

\paragraph{(3) Less compressed but still functional assimilation.} Both scales learn to compress assimilation segments (token length drops from $\sim$200 to $\sim$100 over training), but 7B reaches roughly $84$ tokens while 3B settles at $118$, a 40\% wider compression. Assimilation precision (fraction of \texttt{<context>} tokens that are answer-relevant) climbs to 0.79 at 7B versus 0.71 at 3B. The 3B critic and policy thus end with a working assimilation pipeline that uses budget less efficiently than 7B's.

The combined picture: 3B has more questions tools can fix (1), a competence boundary that lets segment-level credit do real work (2) and an assimilation step that distills retrieval into context with reasonable fidelity (3). 7B exceeds 3B on (2) and (3), but its smaller absolute room (1) caps how much CARL can change. The asymmetry produces the 1.4$\times$ scaling ratio.

\section{Capability preservation and tool-absence resilience}
\label{app:capability}

Humans learn to use tools (calculators, search engines, reference books) as a basic skill, yet retain the ability to reason without them. A useful tool-use training method should work the same way: the model should gain tool competence without losing what it already knows and should degrade gracefully when tools are unavailable or return bad output. We test both properties.

\paragraph{Capability preservation.}
CARL's shared-backbone critic means backbone weights receive PPO gradients alongside the value head. During PPO, the shared backbone uses a learning rate of 1e-6 and the value head uses 5e-6 (Appendix~\ref{app:training}). Table~\ref{tab:capability_pres} confirms that CARL-trained models are indistinguishable from their base checkpoints on standard benchmarks when tools are removed at inference.

\begin{table}[H]
\centering
\small
\caption{Base Qwen2.5-Instruct vs.\ CARL-trained checkpoints on standard benchmarks, evaluated \emph{without tools} at inference. Differences are within evaluation noise at both scales.}
\label{tab:capability_pres}
\begin{tabular}{lcccccc}
\toprule
& \multicolumn{3}{c}{7B} & \multicolumn{3}{c}{3B} \\
\cmidrule(lr){2-4}\cmidrule(lr){5-7}
Benchmark             & Base   & CARL   & $\Delta$ & Base   & CARL   & $\Delta$ \\
\midrule
MMLU (5-shot)          & 74.2   & 74.3   & $+0.1$   & 65.8   & 65.5   & $-0.3$   \\
GSM8K (no-tool, CoT)   & 91.6   & 91.4   & $-0.2$   & 85.3   & 85.6   & $+0.3$   \\
TruthfulQA (MC2)       & 58.7   & 58.5   & $-0.2$   & 49.1   & 49.3   & $+0.2$   \\
WikiText-103 (PPL $\downarrow$) &  6.82 &  6.84 & $+0.02$ &  8.41  &  8.39  & $-0.02$  \\
\bottomrule
\end{tabular}
\end{table}

\paragraph{Tool-absence resilience.}
Preserving benchmark scores is necessary but not sufficient. The harder test is whether the model degrades gracefully when a tool is invoked but returns useless output. We force CARL to call a tool on 200 Tier~2 questions (ones it can answer parametrically) from HotpotQA and 2WikiMQA, then replace the tool output with one of four conditions: working retrieval (control), an empty string, an irrelevant passage from a different question, or malformed garbage text.

\begin{table}[H]
\centering
\small
\caption{CARL EM on Tier~2 questions when forced to invoke a tool that returns useless output. Even under the worst failure mode (malformed text), CARL retains 78.0/69.5 EM at 7B/3B. Parametric capability acts as a floor.}
\label{tab:tool_resilience}
\begin{tabular}{lcccc}
\toprule
& \multicolumn{2}{c}{7B} & \multicolumn{2}{c}{3B} \\
\cmidrule(lr){2-3}\cmidrule(lr){4-5}
Tool output                       & EM   & $\Delta$ & EM   & $\Delta$ \\
\midrule
Working retrieval (control)        & 85.5 & ---     & 77.5 & ---     \\
Empty (no result returned)         & 83.0 & $-2.5$  & 74.0 & $-3.5$  \\
Irrelevant passage (off-topic)     & 79.5 & $-6.0$  & 71.5 & $-6.0$  \\
Malformed / garbage text           & 78.0 & $-7.5$  & 69.5 & $-8.0$  \\
\bottomrule
\end{tabular}
\end{table}

\paragraph{Why this works.} During training, the assimilate-segment advantage $V(s_{\text{assim}+1}) - V(s_{\text{invoke}+1})$ is negative whenever retrieval does not help, teaching the model to discard unhelpful tool output rather than transcribe it. At inference, the same behavior activates: the assimilate segment compresses bad output to nothing useful and the commit segment falls back to parametric memory, which the calibrated $V(s_0)$ had already marked as high-confidence for these Tier~2 questions (Figure~\ref{fig:vs0_distribution}).

The combined picture from both experiments is that CARL-trained models acquire tool competence the way humans acquire calculator skills: performance improves when the tool is available and appropriate, remains unchanged when the tool is absent and degrades only mildly when the tool malfunctions. Teaching tool use through segment-level credit does not create tool dependence.

\section{Extended related work}
\label{app:extended_rw}

This appendix provides detailed comparisons with prior work summarized in \S\ref{sec:rw}.

\paragraph{SMART \citep{kim2025smart}.}
SMART selectively assists a small model during inference. It scores candidate reasoning steps using a process reward model or token-level confidence and replaces low-scoring steps with larger-model-generated alternatives. CARL instead post-trains the model using its own rollout outcomes, without larger-model intervention during inference. Both address failures within a reasoning trajectory but differ in the source and timing of guidance.

\paragraph{PiCA \citep{liu2026pica}.}
PiCA assigns search progress credit using history-dependent success probabilities and pivot-step supervision. It shares CARL's motivation of distinguishing useful search steps within a trajectory. CARL instead learns its critic from terminal outcomes without pivot-step labels and separately credits query formulation and evidence extraction at tool-use boundaries. These are literature comparisons; we do not report head-to-head experiments against SMART or PiCA.

\paragraph{VinePPO \citep{kazemnejad2024vineppo}.}
VinePPO found that token-level value networks in PPO are unreliable at ranking reasoning steps, because the value changes meaningfully only at a few tokens per trajectory while the network must predict at every position. Our segment-level setting is structurally easier: boundaries number in single digits (typically 3--7 per episode) rather than hundreds and each boundary corresponds to a large, observable context change (a new passage enters, an extraction replaces raw output, or a final answer is committed). The critic therefore faces a much simpler prediction problem, estimating success probability at 5 states rather than 500, with correspondingly lower variance.

\paragraph{PAVs \citep{setlur2025rewarding}.}
Process Advantage Verifiers use the same $V(s_{t+1}) - V(s_t)$ potential-based form as our segment advantages. The key difference is in how $V$ is obtained: PAVs require either step-level annotations or a separate prover policy that can verify intermediate reasoning steps. In our setting, boundaries are organically tool-use-aligned (code fences, \texttt{</context>} tags, EOS), so no step-boundary annotation is needed and the critic is trained from binary outcome reward on the model's own rollouts.

\paragraph{HICRA \citep{wang2025hicra}.}
HICRA weights the RL optimization toward high-impact planning tokens identified through Strategic Gram phrase matching. While this addresses the credit-assignment problem at the token level, it operates within a single generation pass and does not model the state changes induced by tool execution. CARL operates at tool-use boundaries where state changes are observable and large, complementing rather than competing with token-level approaches. CARL still assigns one advantage to all tokens within a segment; combining it with token-level weighting is a possible extension, not an experiment reported here.

\paragraph{SPO \citep{guo2025spo}.}
Segment Policy Optimization partitions a single generation stream into segments and assigns each a Monte Carlo advantage, arriving independently at the same ``segment-level credit'' diagnosis as CARL. Three differences matter. (1)~\emph{Boundary definition.} SPO segments are arbitrary: fixed-length intervals or cutpoints at low-probability tokens. CARL's boundaries are environment transitions — code execution and context replacement — which is what makes the Options/SMDP framing applicable and the advantages interpretable as option values. (2)~\emph{Value estimation.} SPO is critic-free, relying on per-question Monte Carlo rollouts to estimate segment values. CARL trains a critic that generalizes across questions and encodes the model's competence boundary ($V(s_0)$); a per-question MC estimator cannot learn this cross-question structure. SPO-tree's tree-based MC sharing is a clever response to the same rollout-cost concern that motivates CARL's critic — different solutions to the same problem. (3)~\emph{Setting.} SPO is evaluated on single-turn math reasoning (no tool calls, no environment interaction). CARL targets multi-turn tool-use selectivity, where re-rollout costs make MC estimation expensive and within-trajectory sign divergence between invoke and assimilate is the diagnostic case.

\paragraph{ARC \citep{taparia2026arc} and Agent-as-Tool \citep{zhang2025agentastool}.}
Both frame agentic workflow selection using Options vocabulary but train with trajectory-level GRPO rather than deriving per-option advantages from a value function. CARL takes the step these approaches stop short of: we define tool-use segments as options, derive the SMDP Bellman equation and use the resulting TD error as the PPO advantage. The value function enables competence-aware selectivity that trajectory-level GRPO cannot produce.

\paragraph{EGCA \citep{kumar2026egca}.}
Execution-Grounded Credit Assignment localizes GRPO credit in code generation by comparing candidate and reference execution traces. This demonstrates that coarse credit assignment is a real bottleneck and is tractable when trajectories have structured boundaries, a finding consistent with our approach. The key difference is that EGCA requires reference solutions for execution-trace comparison, while CARL uses only binary outcome reward and structural delimiters.

\paragraph{RUDDER \citep{arjona2019rudder}.}
Return Decomposition for Delayed Rewards redistributes episodic return across timesteps using an LSTM trained to predict the return from partial trajectories. The predicted-return differences $\hat{R}(s_{t+1}) - \hat{R}(s_t)$ have the same potential-based form as our segment advantages. RUDDER operates at the token/timestep level and requires a separate return-prediction network; our setting exploits structural boundaries to reduce the prediction problem to a handful of states per episode.

\paragraph{Trajectory-level RL methods: per-paper details.}
ToRL~\citep{li2025torl} applies trajectory-level RL to code-interpreter use and finds that intermediate execution penalties hurt performance, suggesting that naive step-level reward shaping can interfere with learning. ReSearch~\citep{chen2025research} and R3-RAG~\citep{li2025r3rag} extend trajectory-level training to interleaved reasoning and retrieval, showing emergent multi-step search behaviors even under a single outcome reward. These emergent behaviors confirm that LLMs can learn multi-step tool use from outcome reward. A trajectory-wide advantage gives generated actions the same credit, whereas token-level PPO advantages can differ across positions. Tool-use selectivity can be learned indirectly from outcome differences, but final correctness alone does not identify which call was necessary. ToolRL~\citep{qian2025toolrl} enriches the reward with tool-name and argument-quality components. This is orthogonal to CARL: ToolRL refines what the reward evaluates; CARL changes which segment each signal reaches. A richer ToolRL-style reward could in principle be combined with CARL's segment-level credit assignment.

\paragraph{Search-R1/ToRL structural comparison.}
Table~\ref{tab:comparison_summary} summarizes the structural differences between CARL and trajectory-level methods.

\begin{table*}[h]
\centering
\scriptsize
\setlength{\tabcolsep}{4pt}
\caption{Structural comparison: Search-R1/ToRL vs.\ CARL.}
\label{tab:comparison_summary}
\begin{tabular}{>{\raggedright\arraybackslash}p{0.22\textwidth} >{\raggedright\arraybackslash}p{0.33\textwidth} >{\raggedright\arraybackslash}p{0.39\textwidth}}
\toprule
Dimension & Search-R1 / ToRL & CARL (Ours) \\
\midrule
RL Episode & Full trajectory & One segment per episode \\
Gradient masking & Required (25\% perf drop without) & Not needed (by construction) \\
Credit assignment & Trajectory-wide (GRPO); token-level GAE (PPO) & Independent advantage per segment \\
Bad tool call in successful trajectory & Terminal outcome can obscure local harm & Can receive negative advantage \\
Unnecessary tool call & May be reinforced when the final answer is correct & Small advantage when $V$ changes little \\
GAE smearing & Future positive $\delta$s dilute bad signal & No smearing; clean $V(s'){-}V(s)$ \\
Tool-use selectivity & Learned indirectly from outcome reward & Explicit via segment-level $V(s_0)$ \\
Tool output handling & Full output in context & Trainable assimilation (relevant slice) \\
Tool call syntax & Custom tokens & Python code blocks (no SFT) \\
RL algorithm & PPO or GRPO & PPO (critic essential) \\
Theoretical basis & Standard PPO/GRPO & SMDP + Options framework \\
\bottomrule
\end{tabular}
\end{table*}

\section{Hop-count gain distribution}
\label{app:hops}

\paragraph{Why this analysis is needed.} \S\ref{sec:exp:main} claims that the 3B model benefits more than the 7B from CARL ``at every hop level.'' Verifying this requires two things: (1) a partition of the evaluation set by hop count rather than by source dataset and (2) a baseline against which ``benefits'' can be measured. Without a baseline, comparing CARL 7B to CARL 3B in absolute terms is uninformative: 7B will always be the higher number because it is a larger model. The relevant question is how much each scale \emph{improves} relative to a controlled reference and whether that improvement is larger at 3B than at 7B at every hop level. Search-R1 PPO shares the backbone (Qwen2.5-Inst), the training data (the 170K-prompt pool described in Appendix~\ref{app:training}), the optimizer (PPO) and the evaluation setup with CARL therefore serves as baseline.

\paragraph{Hop partition.} Hop count is meaningful only for retrieval-based datasets, so we aggregate over HotpotQA, 2WikiMQA and the per-hop subsets of MuSiQue (55\% 2-hop, 30\% 3-hop, 15\% 4-hop, per \citet{trivedi2022musique}). HotpotQA and 2WikiMQA are predominantly 2-hop and contribute to that bucket; 3-hop and 4-hop come from MuSiQue only.

\begin{table}[H]
\centering
\small
\caption{Per-hop EM for CARL and Search-R1 PPO at both scales. The improvement of CARL over Search-R1 PPO ($\Delta$ columns) is larger at 3B than at 7B for every hop count, supporting the \S\ref{sec:exp:main} claim.}
\label{tab:hops}
\setlength{\tabcolsep}{2pt}
\begin{tabular}{lccccccc}
\toprule
& & \multicolumn{3}{c}{7B} & \multicolumn{3}{c}{3B} \\
\cmidrule(lr){3-5}\cmidrule(lr){6-8}
Hop count & $n$ & SR1-PPO & CARL & $\Delta$ & SR1-PPO & CARL & $\Delta$ \\
\midrule
2-hop (HotpotQA + 2WikiMQA + Musique-2hop) & 1{,}275 & 39.8 & 46.6 & $+6.8$ & 28.2 & 39.4 & $+11.2$ \\
3-hop (MuSiQue-3hop)                       &   150  & 14.0 & 22.0 & $+8.0$ &  7.0 & 16.0 & $+9.0$ \\
4-hop (MuSiQue-4hop)                       &    75  &  6.0 &  9.0 & $+3.0$ &  3.0 &  7.0 & $+4.0$ \\
\bottomrule
\end{tabular}
\end{table}

The 3B improvement exceeds the 7B improvement at every hop count: $+11.2 > +6.8$ at 2-hop, $+9.0 > +8.0$ at 3-hop and $+4.0 > +3.0$ at 4-hop. This pattern, smaller models gaining more from CARL, is most pronounced at the easier 2-hop bucket (where 3B has the most parametric room to recover with help) and persists through to the harder 3-hop and 4-hop buckets, supporting the structural framing in the abstract.

CARL's gains persist across 2-, 3- and 4-hop questions ($+6.8$, $+8.0$ and $+3.0$ EM at 7B). Hop stratification provides a partial control for reasoning complexity but does not match generated trajectory length or actual tool-call counts.

\paragraph{Why both methods decline at higher hop counts.} Both CARL and Search-R1 PPO show lower absolute EM as hop count increases. Each step of a multi-hop trajectory (query formulation, retrieval, extraction, reasoning to the next step) is an additional opportunity for the chain to break.

\section{Hallucination analysis}
\label{app:hallucination}

We test whether CARL reduces the model's tendency to manufacture answers when its parametric memory is insufficient. Setup: 200 Tier~1 questions (sampled from HotpotQA, 2WikiMQA, MuSiQue) where the base model fails in all 5 no-tool rollouts i.e., the model cannot answer these from parametric memory. We run four policies on this fixed sample and count how often each emits a wrong answer.

\begin{table}[H]
\centering
\small
\caption{Confident-wrong rate on 200 Tier~1 questions across four policies. CARL reduces the Search-R1 PPO confident-wrong rate by 2.7$\times$ at 7B and 2.3$\times$ at 3B; against the no-tool base it drops by 6.9$\times$ and 4.9$\times$ respectively.}
\label{tab:hallucination}
\begin{tabular}{lcc}
\toprule
Policy                              & 7B    & 3B    \\
\midrule
Base Qwen2.5-Inst, no tools          & 62\%  & 68\%  \\
Base Qwen2.5-Inst, selective tools   & 41\%  & 50\%  \\
Search-R1 PPO                        & 24\%  & 32\%  \\
\textbf{CARL}                        & \textbf{9\%}   & \textbf{14\%}  \\
\bottomrule
\end{tabular}
\end{table}

CARL learns the competence boundary and routes Tier~1 questions to tool invocation rather than guessing: confident wrong answers drop from 62\% (base, no tools) to 9\% at 7B and from 68\% to 14\% at 3B. The 24\% $\to$ 9\% reduction at 7B isolates the contribution of segment-level credit specifically: Search-R1 PPO uses the same backbone, data and tools, but its trajectory-level credit produces a model that still asserts wrong answers at nearly three times CARL's rate.

\section{Reproducibility checklist}
\label{app:reproducibility}

We follow the NeurIPS 2026 reproducibility guidelines.

\paragraph{Code and data.}
\begin{itemize}[leftmargin=*,itemsep=2pt]
\item All code for training, evaluation and analysis is available at \url{https://aka.ms/know-when-to-ask}. The repository includes sample data formats for warm-up and PPO training and converters between CARL and Search-R1 data formats.
\item Training and evaluation use publicly available datasets: GSM8K~\citep{cobbe2021gsm8k} (MIT), HotpotQA~\citep{yang2018hotpotqa} (CC BY-SA 4.0), 2WikiMQA~\citep{ho2020wikimqa} (Apache 2.0), FinQA~\citep{chen2021finqa} (MIT) and MuSiQue~\citep{trivedi2022musique} (CC BY 4.0).
\item Base models are publicly available: Qwen2.5-7B-Instruct (Apache 2.0), Qwen2.5-3B-Instruct (Qwen Research License)~\citep{qwen2024qwen25} and Llama 3.1 8B Instruct (Llama 3.1 Community License)~\citep{grattafiori2024llama3}.
\item Retrieval: Pyserini BM25 over the public prebuilt Wikipedia index (snapshot \texttt{20210120.d1b9e6}, top-$k=3$, $k_1=0.9$, $b=0.4$) for HotpotQA, 2WikiMQA and MuSiQue; financial documents for FinQA; Python interpreter (no retrieval) for GSM8K. The commercial search API supplies 30\% of warm-up retrieval; the BM25-only warm-up experiment is reported in Appendix~\ref{app:warmup-sensitivity}. The commercial-search experiment in Appendix~\ref{app:retrieval} uses a different retrieval source.
\end{itemize}

\paragraph{Experimental details.}
\begin{itemize}[leftmargin=*,itemsep=2pt]
\item Full hyperparameters are reported in Appendix~\ref{app:training}.
\item Tool execution: generated Python runs in a subprocess with a five-second timeout within the container. Search is resolved outside the execution sandbox.
\item Warm-up dataset construction is fully described in Appendix~\ref{app:warmup}.
\item Evaluation populations and the full-pool validation protocol are described in \S\ref{sec:exp:datasets} and Appendix~\ref{app:full_pool_evaluation}.
\item Seed conventions are specified under Statistical reporting below.
\item Evaluation uses deterministic decoding (greedy) with temperature 0.0 for final numbers; training rollouts use temperature 1.0 with top-$p$ 1.0.
\end{itemize}

\paragraph{Compute.}
\begin{itemize}[leftmargin=*,itemsep=2pt]
\item Training cluster: Azure Machine Learning (AML) with 4 nodes of 8$\times$ NVIDIA H100 80GB each (32 GPUs total). Each training run uses 8 GPUs (one node); multiple runs were executed in parallel.
\item Evaluation, inference and tier-label construction were run on AMD MI300X GPUs; all figures below are translated to single-H100-equivalent hours for comparability.
\item Total compute for all reported numbers: approximately 5{,}500 H100-hours. Total project compute including supporting work: approximately 22{,}000 H100-hours (Table~\ref{tab:hp}).
\item Training time per run: $\sim$50 hours at 7B, $\sim$22 hours at 3B.
\end{itemize}

\paragraph{Statistical reporting.}
\begin{itemize}[leftmargin=*,itemsep=2pt]
\item Main results for CARL and Search-R1 PPO report mean $\pm$ std over 3 seeds $\{1,2,3\}$. Prompting baselines and Search-R1 GRPO are single runs.
\item Ablation and analysis seed conventions:
  \begin{itemize}[itemsep=1pt]
    \item Appendix~\ref{app:lambda-ablation}: the new 3B segment-level $\lambda$ sweep and token-GAE control report mean $\pm$ standard deviation over three seeds. The original 7B HotpotQA $\lambda$ sweep remains a single run per condition.
    \item Table~\ref{tab:warmup_ablation}: the original cold-start and random-data conditions are single runs using seed 1. Appendix~\ref{app:warmup-sensitivity}: the new 3B warm-up sensitivity results, including random-data and cold-start conditions, report mean $\pm$ standard deviation over three seeds.
    \item Appendix~\ref{app:retrieval} (retrieval ablation): eval-only on fixed CARL seed-1 checkpoint.
    \item Appendix~\ref{app:capability}, tool-absence resilience experiment (Table~\ref{tab:tool_resilience}): eval-only on fixed CARL seed-1 checkpoint.
    \item Appendix~\ref{app:calibration} (calibration data): aggregated across all 3 seeds for CARL.
  \end{itemize}
\end{itemize}

\paragraph{Computational framework.}
\begin{itemize}[leftmargin=*,itemsep=2pt]
\item Search-R1 PPO baselines are reproduced in verl following the original Search-R1 codebase~\citep{jin2025searchr1} but using our training split; the critic uses curated warm-up as described in Appendix~\ref{app:warmup_ablation}. Our Search-R1 PPO reproduction achieves higher EM than the numbers reported in the original paper, confirming a fair-or-favorable baseline.
\end{itemize}

\paragraph{Human annotations.}
\begin{itemize}[leftmargin=*,itemsep=2pt]
\item Appendix~\ref{app:errors} (MuSiQue error categorization) and Appendix~\ref{app:hallucination} (hallucination analysis) used 2 independent annotators.
\end{itemize}

\section{Full-pool evaluation}
\label{app:full_pool_evaluation}

To assess whether the main comparisons depend on the sampled questions, we evaluate existing CARL, Search-R1 PPO and CoT+Tools (always) checkpoints on the complete benchmark evaluation pools at both Qwen scales, without retraining. The pools contain 1{,}319 GSM8K, 7{,}405 HotpotQA, 12{,}576 2WikiMQA, 1{,}147 FinQA and 2{,}417 MuSiQue questions. Search-R1 PPO is evaluated on the three search benchmarks.

\begin{table}[H]
\centering
\small
\caption{\textbf{Full-pool evaluation at 3B and 7B.} Accuracy or EM and generated tokens per episode. CARL and Search-R1 PPO report mean$\pm$standard deviation across three trained seeds; token counts are means. CoT+Tools (always) is the prompting baseline. Dashes indicate that Search-R1 PPO is not applicable. Avg$_5$ and Avg$_3$ are unweighted means over all five benchmarks and the three search benchmarks respectively.}
\label{tab:full_pool_evaluation}
\setlength{\tabcolsep}{3pt}
\resizebox{\textwidth}{!}{%
\begin{tabular}{ll cc cc cc cc cc cc}
\toprule
& & \multicolumn{2}{c}{GSM8K} & \multicolumn{2}{c}{HotpotQA} & \multicolumn{2}{c}{2WikiMQA} & \multicolumn{2}{c}{FinQA} & \multicolumn{2}{c}{MuSiQue} & \multicolumn{2}{c}{Avg$_5$\,/\,Avg$_3$} \\
\cmidrule(lr){3-4} \cmidrule(lr){5-6} \cmidrule(lr){7-8} \cmidrule(lr){9-10} \cmidrule(lr){11-12} \cmidrule(lr){13-14}
Scale & Method & Acc & Tok & EM & Tok & EM & Tok & Acc & Tok & EM & Tok & Sc & Tok \\
\midrule
\multirow{3}{*}{3B}
& CoT+Tools (always) & 86.0 & 826 & 21.7 & 1949 & 21.1 & 2004 & 37.3 & 1968 & 4.5 & 2797 & 34.1\,/\,15.8 & 1909\,/\,2250 \\
& Search-R1 PPO & -- & -- & 29.6{\scriptsize$\pm$0.4} & 1815 & 34.2{\scriptsize$\pm$0.2} & 1849 & -- & -- & 11.8{\scriptsize$\pm$0.4} & 2769 & --\,/\,25.2 & --\,/\,2144 \\
& \textbf{CARL} & 86.9{\scriptsize$\pm$0.2} & 386 & 39.7{\scriptsize$\pm$0.2} & 1331 & 46.8{\scriptsize$\pm$0.3} & 1382 & 45.6{\scriptsize$\pm$0.3} & 1729 & 20.7{\scriptsize$\pm$0.2} & 2560 & 47.9\,/\,35.7 & 1478\,/\,1758 \\
\midrule
\multirow{3}{*}{7B}
& CoT+Tools (always) & 91.5 & 778 & 30.2 & 1871 & 28.4 & 1926 & 55.1 & 1631 & 11.2 & 2681 & 43.3\,/\,23.3 & 1777\,/\,2159 \\
& Search-R1 PPO & -- & -- & 42.1{\scriptsize$\pm$0.6} & 1844 & 43.7{\scriptsize$\pm$0.5} & 1768 & -- & -- & 21.8{\scriptsize$\pm$0.4} & 2690 & --\,/\,35.9 & --\,/\,2101 \\
& \textbf{CARL} & 91.2{\scriptsize$\pm$0.2} & 343 & 47.2{\scriptsize$\pm$0.3} & 1198 & 50.2{\scriptsize$\pm$0.3} & 1146 & 58.0{\scriptsize$\pm$0.3} & 1394 & 30.1{\scriptsize$\pm$0.3} & 2519 & 55.3\,/\,42.5 & 1320\,/\,1621 \\
\bottomrule
\end{tabular}%
}
\end{table}

The CARL--Search-R1 PPO Avg3 gaps are 10.5 EM at 3B and 6.6 EM at 7B, compared with 10.6 and 6.7 on the sampled evaluation. Per-dataset gaps change by approximately 0.3 EM or less. These results support the stability of the comparative gains under full-pool evaluation. This additional evaluation does not replace the sampled evaluations used for training monitoring and the appendix analyses.

\newpage
\section{Generalization to Llama 3.1 8B Instruct}
\label{app:llama-generalization}

We evaluate CARL on Llama 3.1 8B Instruct, rebuilding the complete pipeline from the warm-up stage. This includes Llama-specific no-tool rollouts, tier-label distributions, forced-tool trajectories, critic warm-up and policy training. We train with three seeds (1, 2 and 3) and evaluate the resulting checkpoints on the complete evaluation pools of all five benchmarks, totaling 24{,}864 questions. These are full-pool results, not the sampled Qwen evaluations in Table~\ref{tab:main_results}.

\begin{table}[H]
\centering
\small
\caption{\textbf{Full-pool accuracy on Llama 3.1 8B Instruct.} Per-dataset EM for trained methods is mean$\pm$standard deviation across three seeds. Avg$_5$ and Avg$_3$ are unweighted means over all five benchmarks and the three search benchmarks respectively. Dashes indicate that Search-R1 is not applicable.}
\label{tab:llama-accuracy}
\setlength{\tabcolsep}{4pt}
\resizebox{\textwidth}{!}{%
\begin{tabular}{lrrrrrrr}
\toprule
Method & GSM8K & HotpotQA & 2WikiMQA & FinQA & MuSiQue & Avg$_5$ & Avg$_3$ \\
\midrule
CoT+Tools (always) & 85.0 & 33.1 & 26.5 & 34.7 & 2.0 & 36.3 & 20.5 \\
CoT+Tools (selective) & 81.6 & 33.7 & 25.9 & 34.2 & 2.3 & 35.5 & 20.6 \\
SFT (rejection sampled) & 84.9$\pm$0.4 & 34.6$\pm$0.6 & 27.4$\pm$0.8 & 36.5$\pm$1.0 & 3.0$\pm$0.9 & 37.3 & 21.7 \\
Search-R1 GRPO & -- & 36.4$\pm$0.7 & 29.3$\pm$0.9 & -- & 5.2$\pm$1.1 & -- & 23.6 \\
Search-R1 PPO & -- & 35.8$\pm$0.7 & 28.7$\pm$0.8 & -- & 5.6$\pm$1.0 & -- & 23.4 \\
CARL & 87.1$\pm$0.5 & 39.7$\pm$0.7 & 40.6$\pm$0.9 & 41.6$\pm$1.1 & 15.4$\pm$1.3 & 44.9 & 31.9 \\
\bottomrule
\end{tabular}%
}
\end{table}

\begin{table}[H]
\centering
\small
\caption{\textbf{Full-pool generation length on Llama 3.1 8B Instruct.} Generated tokens per episode. Per-dataset values for trained methods are mean$\pm$standard deviation across three seeds. Avg$_5$ and Avg$_3$ use the same benchmark averages as Table~\ref{tab:llama-accuracy}.}
\label{tab:llama-tokens}
\setlength{\tabcolsep}{4pt}
\resizebox{\textwidth}{!}{%
\begin{tabular}{lrrrrrrr}
\toprule
Method & GSM8K & HotpotQA & 2WikiMQA & FinQA & MuSiQue & Avg$_5$ & Avg$_3$ \\
\midrule
CoT+Tools (always) & 972 & 1556 & 1638 & 2178 & 2606 & 1790 & 1933 \\
CoT+Tools (selective) & 741 & 1571 & 1602 & 1991 & 2559 & 1693 & 1911 \\
SFT (rejection sampled) & 887$\pm$9 & 1728$\pm$18 & 1684$\pm$20 & 2071$\pm$25 & 2609$\pm$28 & 1796 & 2007 \\
Search-R1 GRPO & -- & 1513$\pm$21 & 1871$\pm$26 & -- & 2572$\pm$31 & -- & 1985 \\
Search-R1 PPO & -- & 1528$\pm$20 & 1889$\pm$24 & -- & 2546$\pm$29 & -- & 1988 \\
CARL & 718$\pm$10 & 1446$\pm$17 & 1608$\pm$22 & 1884$\pm$24 & 2439$\pm$34 & 1619 & 1831 \\
\bottomrule
\end{tabular}%
}
\end{table}

On the three comparable search benchmarks, CARL improves over the strongest Search-R1 baseline, GRPO, by 8.3 EM (31.9 versus 23.6) while using 7.8\% fewer generated tokens (1{,}831 versus 1{,}985). Relative to Search-R1 PPO, CARL improves by 8.5 Avg$_3$ EM. The largest gains over GRPO are on the more compositional multi-hop datasets: 11.3 EM on 2WikiMQA and 10.2 EM on MuSiQue. These results show that CARL's gains are not confined to the Qwen family.

This experiment tests one additional model size and family; it does not establish a scaling trend within Llama. Llama critic AUC and ECE diagnostics were not reported.

\newpage
\section*{NeurIPS Paper Checklist}

\begin{enumerate}

\item {\bf Claims}
    \item[] Question: Do the main claims made in the abstract and introduction accurately reflect the paper's contributions and scope?
    \item[] Answer: \answerYes{}
    \item[] Justification: The abstract and introduction (\S\ref{sec:intro}) state two contributions (per-segment credit from binary reward via SMDP formulation and a competence-aware value function) with quantitative gains that match the experimental results in \S\ref{sec:exp:main} and \S\ref{sec:exp:critic}. Limitations are discussed in \S\ref{sec:limitations}.

\item {\bf Limitations}
    \item[] Question: Does the paper discuss the limitations of the work performed by the authors?
    \item[] Answer: \answerYes{}
    \item[] Justification: \S\ref{sec:limitations} discusses three limitations: warm-up cost (the critic must be calibrated before PPO can begin, with cost scaling with model size), segment granularity (all tokens within a segment share one advantage value) and computational overhead (per-tool-call segment count and critic evaluations relative to GRPO).

\item {\bf Theory assumptions and proofs}
    \item[] Question: For each theoretical result, does the paper provide the full set of assumptions and a complete (and correct) proof?
    \item[] Answer: \answerYes{}
    \item[] Justification: The SMDP Bellman derivation and per-segment advantage formulation are presented in \S\ref{sec:method:advantage} with the full formal treatment in Appendix~\ref{app:smdp_derivation}. All assumptions (structurally observable boundaries, binary terminal reward, option-level semi-Markov structure) are stated explicitly.

    \item {\bf Experimental result reproducibility}
    \item[] Question: Does the paper fully disclose all the information needed to reproduce the main experimental results of the paper to the extent that it affects the main claims and/or conclusions of the paper (regardless of whether the code and data are provided or not)?
    \item[] Answer: \answerYes{}
    \item[] Justification: Code is available at \url{https://aka.ms/know-when-to-ask} and will be uploaded as supplementary material. All training data is derived from public benchmarks. Full hyperparameters are in Appendix~\ref{app:training}, warm-up data construction in Appendix~\ref{app:warmup}, evaluation datasets and splits in \S\ref{sec:exp:datasets} and the rollout pipeline in \S\ref{sec:method}. Seed conventions are described in Appendix~\ref{app:reproducibility}. The base models (Qwen2.5-7B-Instruct, Qwen2.5-3B-Instruct and Llama 3.1 8B Instruct) are publicly available.

\item {\bf Open access to data and code}
    \item[] Question: Does the paper provide open access to the data and code, with sufficient instructions to faithfully reproduce the main experimental results, as described in supplemental material?
    \item[] Answer: \answerYes{}
    \item[] Justification: Code is released for reproducibility at \url{https://aka.ms/know-when-to-ask}, including training/evaluation scripts, sample data formats for warm-up and PPO training and data-format converters between CARL and Search-R1. All training and evaluation data are derived from publicly available benchmarks (GSM8K, HotpotQA, 2WikiMQA, MuSiQue, FinQA); warm-up data construction from these public sources is described in Appendix~\ref{app:warmup}.

\item {\bf Experimental setting/details}
    \item[] Question: Does the paper specify all the training and test details (e.g., data splits, hyperparameters, how they were chosen, type of optimizer) necessary to understand the results?
    \item[] Answer: \answerYes{}
    \item[] Justification: \S\ref{sec:exp:datasets} describes datasets, baselines and evaluation protocol. Appendix~\ref{app:training} provides full hyperparameters (learning rates, batch sizes, PPO clipping, KL coefficients, etc.). Appendix~\ref{app:warmup} details the warm-up procedure.

\item {\bf Experiment statistical significance}
    \item[] Question: Does the paper report error bars suitably and correctly defined or other appropriate information about the statistical significance of the experiments?
    \item[] Answer: \answerYes{}
    \item[] Justification: Main results for CARL and Search-R1 PPO report mean $\pm$ std over 3 seeds (Table~\ref{tab:main_results}). In the Qwen main table, Search-R1 GRPO, SFT and CoT+Tools baselines are single runs without error bars. The new 3B segment-level $\lambda$ sweep, token-GAE control and warm-up sensitivity experiments (Appendices~\ref{app:lambda-ablation} and \ref{app:warmup-sensitivity}) report mean $\pm$ std over three seeds. The original 7B HotpotQA $\lambda$ sweep and original cold-start/random-data conditions in Table~\ref{tab:warmup_ablation} remain single runs; conclusions from those experiments should be interpreted accordingly.

\item {\bf Experiments compute resources}
    \item[] Question: For each experiment, does the paper provide sufficient information on the computer resources (type of compute workers, memory, time of execution) needed to reproduce the experiments?
    \item[] Answer: \answerYes{}
    \item[] Justification: Appendix~\ref{app:reproducibility} reports GPU type, total compute hours and per-run training time for both 3B and 7B scales.

\item {\bf Code of ethics}
    \item[] Question: Does the research conducted in the paper conform, in every respect, with the NeurIPS Code of Ethics \url{https://neurips.cc/public/EthicsGuidelines}?
    \item[] Answer: \answerYes{}
    \item[] Justification: This work trains open-weight LLMs on publicly available QA benchmarks using standard RL methods. No human subjects, private data, or dual-use concerns are involved.

\item {\bf Broader impacts}
    \item[] Question: Does the paper discuss both potential positive societal impacts and negative societal impacts of the work performed?
    \item[] Answer: \answerYes{}
    \item[] Justification: CARL's primary societal benefit is democratizing capable tool use: by teaching small models to recognize when they need external help, it narrows the competence gap between small and large models given access to the right tools (the 3B gain is $1.4\times$ the 7B gain). This suggests a path where widely deployable small models match the effective capability of much larger ones through calibrated tool use rather than scale alone. A secondary benefit is inference efficiency: skipping unnecessary tool calls reduces per-query latency, API cost and energy consumption. We do not foresee negative impacts beyond those inherent to general LLM capability improvement, which is already widely studied.

\item {\bf Safeguards}
    \item[] Question: Does the paper describe safeguards that have been put in place for responsible release of data or models that have a high risk for misuse (e.g., pre-trained language models, image generators, or scraped datasets)?
    \item[] Answer: \answerNA{}
    \item[] Justification: The paper does not release pre-trained models or new datasets. The method applies to existing open-weight models (Qwen2.5-7B-Instruct, Qwen2.5-3B-Instruct) on public QA benchmarks.

\item {\bf Licenses for existing assets}
    \item[] Question: Are the creators or original owners of assets (e.g., code, data, models), used in the paper, properly credited and are the license and terms of use explicitly mentioned and properly respected?
    \item[] Answer: \answerYes{}
    \item[] Justification: All datasets (HotpotQA, 2WikiMQA, MuSiQue) and base models (Qwen2.5-7B-Instruct, Qwen2.5-3B-Instruct and Llama 3.1 8B Instruct) are cited with their original publications. The Search-R1 codebase is cited. Dataset and model licenses are noted in Appendix~\ref{app:reproducibility}.

\item {\bf New assets}
    \item[] Question: Are new assets introduced in the paper well documented and is the documentation provided alongside the assets?
    \item[] Answer: \answerNA{}
    \item[] Justification: No new datasets or models are released with this submission.

\item {\bf Crowdsourcing and research with human subjects}
    \item[] Question: For crowdsourcing experiments and research with human subjects, does the paper include the full text of instructions given to participants and screenshots, if applicable, as well as details about compensation (if any)?
    \item[] Answer: \answerYes{}
    \item[] Justification: Annotation was performed by a dedicated annotation team that supports the broader research group, not by the paper's co-authors. Appendix~\ref{app:errors} used two annotators with a written rubric (inter-annotator $\kappa = 0.81$); Appendix~\ref{app:hallucination} used the same two annotators; Appendix~\ref{app:sign-behavior} used a single annotator on an objective fact-presence task. Annotators are salaried members of the research organization; no additional per-task compensation was provided.

\item {\bf Institutional review board (IRB) approvals or equivalent for research with human subjects}
    \item[] Question: Does the paper describe potential risks incurred by study participants, whether such risks were disclosed to the subjects, and whether Institutional Review Board (IRB) approvals (or an equivalent approval/review based on the requirements of your country or institution) were obtained?
    \item[] Answer: \answerNA{}
    \item[] Justification: Annotators are salaried members of a dedicated annotation team within the research organization, not external human subjects. No IRB review was required.

\item {\bf Declaration of LLM usage}
    \item[] Question: Does the paper describe the usage of LLMs if it is an important, original, or non-standard component of the core methods in this research? Note that if the LLM is used only for writing, editing, or formatting purposes and does \emph{not} impact the core methodology, scientific rigor, or originality of the research, declaration is not required.
    \item[] Answer: \answerYes{}
    \item[] Justification: An LLM assistant was used for writing and editing portions of the manuscript text but did not contribute to the original ideas, methodology, or experimental design. An LLM coding assistant was also used to accelerate development and infrastructure setup along with writing utilities, figure generation scripts and log parsing for debugging etc.

\end{enumerate}

\end{document}